\documentclass[journal]{vgtc}                     

\onlineid{0}
\vgtccategory{Research}
\title{A Multi-Attribute Latent Space for Visual Analysis of Watches}

\author{%
  \authororcid{Kai Lawonn}{0000-0002-1511-4022},
  \authororcid{Tobias Günther}{0000-0002-3020-0930}, and 
  \authororcid{Monique Meuschke}{0000-0003-2183-6619}
}
\authorfooter{
  \item
  	Kai Lawonn is with Leipzig University and ScaDS.AI Dresden/Leipzig.
  	E-mail: kai.lawonn@uni-leipzig.de.
  \item
  	Tobias Günther is with Friedrich-Alexander-Universität Erlangen-Nürnberg.
  	E-mail: tobias.guenther@fau.de.

  \item Monique Meuschke is with Otto von Guericke University of Magdeburg.
  	E-mail: meuschke@isg.cs.uni-magdeburg.de.
}

\abstract{%
\revise{%
We present a design rationale, embedding model, and interactive visual-analysis system for exploring large wristwatch collections through heterogeneous visual and semantic attributes. The system addresses a common limitation of catalog and e-commerce interfaces: users can filter by metadata, but they receive little support for open-ended exploration of visual similarity, stylistic alternatives, and mixed aesthetic-functional criteria. We therefore represent watches with separate attribute graphs for dial color and dial design, while using watch type as an explicit semantic organizer. Dials are segmented with a U-Net, watch types are predicted with a Vision Transformer, colors are represented through a shared CIELAB reference palette, and dial structure is described with a gradient-based image descriptor. We extend UMAP by combining attribute-specific neighborhood graphs in a unified probabilistic objective and by adding a class-aware layout term that separates global type structure from local visual neighborhoods. The resulting map is exposed in an interactive interface with spatial navigation, metadata filtering, detail inspection, and search-by-example insertion. We evaluate the approach through parameter analysis, runtime measurements, and a qualitative pilot study with watch experts and novices. The results suggest that the system supports discovery and comparison, while also revealing limitations in scalability assessment, search-by-example validation, and the need for broader domain studies. We explicitly discuss these limitations and derive design implications for multi-attribute latent-space visualization across heterogeneous visual collections.
}
}

\keywords{High-dimensional Data, Dimensionality Reduction}

\graphicspath{{figs/}{figures/}{pictures/}{images/}{./}} 

\usepackage{tabu}                      
\usepackage{booktabs}                  
\usepackage{lipsum}                    
\usepackage{mwe}                       
\usepackage{ccicons}                   
\usepackage{nicefrac}

\usepackage{mathptmx}                  
\usepackage{pgfplots}
\pgfplotsset{compat=1.18}
\usepackage{sansmath} 
\usepackage{adjustbox} 
\usetikzlibrary{arrows.meta, positioning, calc}
\usepackage{amsmath}
\usepackage{enumitem}
\usepackage{todonotes}
\newcommand{\R}{\mathrm  I\!\mathrm R}

\newcommand{\vh}{\mathbf{h}}

\newcommand{\vx}{\mathbf{x}}
\newcommand{\vy}{\mathbf{y}}

\newcommand{\cF}{\mathcal{F}}

\DeclareRobustCommand{\rchi}{{\mathpalette\irchi\relax}}
\newcommand{\irchi}[2]{\raisebox{\depth}{$#1\chi$}}

\usepackage{comment}

\makeatletter
\newcommand{\customlabel}[2]{%
   \protected@write \@auxout {}{\string \newlabel {#1}{{#2}{\thepage}{#2}{#1}{}} }%
   \hypertarget{#1}{#2}
}
\makeatother

\usepackage{tabu}                      
\usepackage{booktabs}                  
\usepackage{lipsum}                    
\usepackage{mwe}                       
\usepackage{ccicons}                   

\usepackage{mathptmx}                  
\usepackage{multirow}
\usepackage{tabu}                      
\usepackage{booktabs}                  
\usepackage{lipsum}                    
\usepackage{mwe}                       
\usepackage{todonotes}
\usepackage{mathptmx}                  
\usepackage{pgfplots}
\pgfplotsset{compat=1.18}
\usepackage{sansmath} 
\usepackage{adjustbox} 
\usetikzlibrary{arrows.meta, positioning, calc}
\usepackage{amsmath}
\usepackage{enumitem}

\usepackage{xcolor}
\usepackage{xparse}

\definecolor{revisioncolor}{RGB}{0,0,0}
\NewDocumentCommand{\revise}{+m}{{\color{revisioncolor}#1}}

\makeatletter
\AtBeginDocument{%
  \@ifpackageloaded{hyperref}{\pdfstringdefDisableCommands{\def\revise#1{#1}}}{}%
}
\makeatother

\usepackage{tikz}

\definecolor{likert1}{RGB}{202,0,32}    
\definecolor{likert2}{RGB}{244,165,130} 
\definecolor{likert3}{RGB}{247,247,247} 
\definecolor{likert4}{RGB}{146,197,222} 
\definecolor{likert5}{RGB}{5,113,176}   

\definecolor{likert71}{RGB}{178,24,43}    
\definecolor{likert72}{RGB}{239,138,98} 
\definecolor{likert73}{RGB}{253,219,199} 
\definecolor{likert74}{RGB}{247,247,247} 
\definecolor{likert75}{RGB}{209,229,240}   
\definecolor{likert76}{RGB}{103,169,207}   
\definecolor{likert77}{RGB}{33,102,172}   

\newcommand{\bigdot}[2][2mm]{%
    \tikz \fill[#2] (0,0) circle (#1);%
}

\newcommand{\likert}{
    \[
    \begin{array}{>{\centering\arraybackslash}p{0.15\linewidth}
                    >{\centering\arraybackslash}p{0.15\linewidth}
                    >{\centering\arraybackslash}p{0.15\linewidth}
                    >{\centering\arraybackslash}p{0.15\linewidth}
                    >{\centering\arraybackslash}p{0.15\linewidth}}
        \bigdot[2mm]{likert1} &
        \bigdot[2mm]{likert2} &
        \bigdot[2mm]{likert3} &
        \bigdot[2mm]{likert4} &
        \bigdot[2mm]{likert5} \\
        \text{1} &
        \text{2} &
        \text{3} &
        \text{4} &
        \text{5}
    \end{array}
    \]
}

\newcommand{\likertSeven}{
    \[
    \begin{array}{>{\centering\arraybackslash}p{0.1\linewidth}
                    >{\centering\arraybackslash}p{0.1\linewidth}
                    >{\centering\arraybackslash}p{0.1\linewidth}
                    >{\centering\arraybackslash}p{0.1\linewidth}
                    >{\centering\arraybackslash}p{0.1\linewidth}
                    >{\centering\arraybackslash}p{0.1\linewidth}
                    >{\centering\arraybackslash}p{0.1\linewidth}}
        \bigdot[2mm]{likert71} &
        \bigdot[2mm]{likert72} &
        \bigdot[2mm]{likert73} &
        \bigdot[2mm]{likert74} &
        \bigdot[2mm]{likert75} &
        \bigdot[2mm]{likert76} &
        \bigdot[2mm]{likert77} \\
        \text{-3} &
        \text{-2} &
        \text{-1} &
        \text{0} &
        \text{1} &
        \text{2} &
        \text{3} 
    \end{array}
    \]
}

\newcommand{\freeresponse}[1][0.9\linewidth]{%
    \rule{#1}{0.5pt}%
}

\AtBeginDocument{

}

\begin{document}

\firstsection{Introduction}

\maketitle

\revise{%
The design space of wristwatches spans functional categories (e.g., chronographs, divers, and dress watches), aesthetic characteristics (e.g., dial color, texture, indices, hands, and subdials), and practical constraints (e.g., brand, price, case size, and materials). This mixture makes exploring watches difficult for two distinct user groups. Enthusiasts and collectors often have domain-specific expectations and compare small visual differences across models, while casual users may begin with vague preferences such as ``a blue diver that looks clean'' or ``something similar to this reference watch''. Conventional catalog interfaces support precise metadata filtering, but they do not provide an overview of visual design relationships or a way to move progressively from broad categories to visually similar alternatives.
}

\revise{%
We use a latent-space map as a complementary visual abstraction rather than a replacement for faceted search. A map is useful when users do not yet know the exact combination of filters they need: they can first orient themselves in the collection, identify visually coherent regions, and then inspect or filter nearby items. At the same time, a simple projection of all available features is problematic because watch type, dial color, dial structure, and metadata have different semantics, dimensionalities, and statistical distributions. A concatenated feature vector can make one attribute dominate the distance metric, while a single learned embedding can hide which attribute is responsible for a neighborhood.
}

\revise{%
We therefore separate three roles in the representation. Watch type provides a global semantic scaffold; color and design define local similarity neighborhoods; and metadata such as price or case size remains available as interactive filters. To construct the attributes, we analyze approximately 33,000 watch images. We segment the dial with a U-Net, classify the watch type with a Vision Transformer (ViT), describe color with a shared CIELAB reference palette, and describe dial structure with Histograms of Oriented Gradients (HOG). The resulting representation is designed for exploration tasks in which users compare local alternatives, discover related watches across brands, and insert a reference image as a query.
}

\revise{%
Our embedding builds on UMAP because UMAP explicitly represents high-dimensional neighborhoods as fuzzy graphs and optimizes a corresponding low-dimensional graph. This graph formulation provides a direct way to construct separate neighborhood graphs for heterogeneous attributes and combine them using user-controllable weights. We further add membership-constrained neighborhoods and a decaying class-aware layout term that uses watch type to stabilize the global arrangement without fully overriding local visual similarity. This addresses a core design trade-off: a useful overview should reveal major watch types, but the distances inside each region should still reflect visual attributes such as color and dial design.
}

\revise{%
Building on the embedding, we design an interactive visualization system with spatial navigation, hover-based detail inspection, metadata filtering, and search-by-example insertion. The interface supports a progressive workflow: users start with an overview, narrow exploration to a type or visual region, compare nearby models, and optionally apply practical constraints. We evaluate the system through parameter analysis, runtime measurements, and a qualitative pilot study with experts and novices. Because the number of participants is small, we treat the results as evidence about usage patterns and design implications rather than as a statistical proof of effectiveness.
}
Our main contributions are as follows:
\begin{itemize}[noitemsep, topsep=2pt]
    \item \revise{We formulate a multi-attribute UMAP variant that combines attribute-specific fuzzy neighborhood graphs for heterogeneous visual descriptors and augments them with membership-constrained neighborhoods and a decaying class-aware layout term.}
    \item \revise{We present a review-driven design rationale for a wristwatch exploration system that connects domain requirements to latent-space mapping, metadata filtering, and search-by-example interactions.}
    \item \revise{We construct a wristwatch latent space from segmented dial images, watch-type predictions, color descriptors, and design descriptors, and expose it through an interactive visual-analysis interface.}
    \item \revise{We report a qualitative pilot evaluation with experts and novices, discuss the limits of the current evidence, and derive design implications for visualizing multi-attribute latent spaces in heterogeneous visual collections.}
\end{itemize}

\newpage

\section{Related Work}
\label{Sec:RelatedWork}
\revise{%
We position our work with respect to visual exploration of image collections, high-dimensional visual analytics, and non-linear dimensionality reduction. The review is organized around the design decisions that motivated our system: why a spatial latent map is useful, why it should be combined with coordinated filters and details, and why UMAP is an appropriate foundation for a multi-attribute extension.
}

\subsection{\revise{Visual Exploration of Image and Product Collections}}
\revise{%
Large image collections are often explored through grids, ranked lists, or search interfaces. These designs are effective when the query is known in advance, but they provide limited support for discovering neighborhoods, alternatives, and transitions between visual styles. Visualization research has therefore explored spatial image maps and progressive navigation for large-scale image exploration~\cite{Pezzotti2017}, as well as latent-space cartography for interpreting vector-space embeddings~\cite{Liu2019LatentSpaceCartography}. In product domains, learned visual similarity has been used to compare and retrieve furniture and fashion items~\cite{Bell2015,Han2017}. These systems motivate our use of spatial neighborhoods for discovery, but our setting differs in two ways: the collection combines semantic type labels with multiple visual attributes, and users need to understand which attribute drives a local neighborhood.
}
\revise{%
Recent visual analytics systems have further investigated how large visual collections can be organized beyond plain grids or isolated scatterplots. DendroMap~\cite{Bertucci2023DendroMap} adapts treemap layouts to hierarchically clustered image representations, thereby supporting overview, zooming, and inspection of large image datasets. VISAtlas~\cite{Ye2024VISAtlas} uses neural image embeddings to support exploration and image-based retrieval in large visualization collections. Related systems also use visual analytics to inspect learned representations and model behavior. DeepVID~\cite{Wang2019DeepVID} explains image classifiers through locally faithful surrogate models, while VATLD~\cite{Gou2021VATLD} combines representation learning and visual analytics to assess and improve traffic-light detectors. These works motivate our use of visual overviews, image neighborhoods, and coordinated inspection; however, our goal is not primarily dataset diagnosis or model debugging. Instead, we construct a product-oriented latent map in which different visual and semantic attributes can be explicitly balanced during embedding. }

\revise{%
A map alone is not sufficient for product exploration. Users also need metadata filters, detail views, and task-specific entry points. We therefore combine the embedding with coordinated product metadata and search-by-example insertion. This design choice follows a visual analytics perspective: the latent space provides an overview and neighborhood structure, while filters and details support targeted decisions and reduce the risk of over-interpreting distances.
}

\subsection{\revise{High-Dimensional Data Visualization and Design Alternatives}}
\revise{%
High-dimensional visualization includes data transformation, visual mapping, and view transformation~\cite{Liu2017}. Classical techniques such as scatterplot matrices and parallel coordinates~\cite{Mordechai2010,HeinrichWeiskopf2013} explicitly expose dimensions, but they scale poorly for image collections with thousands of items and do not directly show image-neighborhood relationships. Rank-by-feature approaches~\cite{SeoShneiderman2004}, class-consistency measures~\cite{Sips2009}, and automated view recommendations~\cite{Tatu2009} help users select informative projections, while quality metrics can support the evaluation of projection results~\cite{Bertini2011}. These ideas inform our parameter analysis and our use of class-consistency-style neighborhood reporting in \autoref{Sec:ParameterEvaluation}.
}

\revise{%
Alternative layouts are possible. Faceted grids make metadata categories explicit; cluster-based layouts emphasize group membership; and coordinated attribute views show individual dimensions more transparently. Cluster-aware grid layouts~\cite{Zhou2024ClusterAwareGrid} further show that grid arrangements can be optimized to preserve proximity, compactness, and convexity of cluster structures. Such layouts are attractive when discrete item placement and cluster legibility are the primary goals. We chose a latent-space map because our target tasks involve open-ended exploration of visual similarity and discovery across brands, not only filtering by known attributes or presenting compact clusters. The interface keeps filtering and details available to compensate for the ambiguity of projected distances. }

\subsection{Non-Linear Dimensionality Reduction}
\paragraph*{\revise{t-SNE and Related Methods.}}
\revise{%
The t-distributed Stochastic Neighbor Embedding (t-SNE)~\cite{Maaten2008:tSNE} models local relationships with Gaussian distributions in high-dimensional space and Student's $t$-distributions in low-dimensional space. Perplexity controls the effective neighborhood size~\cite{wattenberg2016how}, and GPU implementations improve scalability~\cite{Chan2018,Pezzotti20}. Recent work also incorporates classifier information into projections, for example, by jointly embedding features and class-probability distributions~\cite{Meng2024}. Such methods are valuable for local cluster analysis, but the optimization does not expose the same fuzzy graph structure that we use to combine multiple attribute-specific neighborhoods.
Regardless, we later examine a generalization of our formulation to t-SNE and observe that the UMAP variant performs best for most mixed-attribute settings, while the t-SNE variant remains competitive.
}

\paragraph*{UMAP.}
\revise{%
UMAP~\cite{McInnes2020} constructs a weighted graph representation of the data manifold and optimizes a low-dimensional graph to match it. For each pair of points $(i,j)$, UMAP estimates the directed membership strength
}
\begin{align}\label{Eq:MembershipProbability}
\tilde p_{ij}=\exp\left(-\frac{\max(0,d(\vx_i,\vx_j)-\rho_i)}{\sigma_i}\right),
\end{align}
\revise{%
where $d(\vx_i,\vx_j)$ is the high-dimensional distance, $\rho_i$ controls local connectivity, and $\sigma_i$ normalizes the local neighborhood scale. Directed memberships are symmetrized as $p_{ij}=\tilde p_{ij}+\tilde p_{ji}-\tilde p_{ij}\tilde p_{ji}$. In the low-dimensional space, UMAP defines
}
\begin{align}
q_{ij}=\frac{1}{1+a\lVert\vy_i-\vy_j\rVert^{2b}},
\end{align}
\revise{%
where $a$ and $b$ control the curve shape. The embedding minimizes the cross-entropy between the high- and low-dimensional fuzzy sets:
}
\begin{align}\label{Eq:CrossEntropy}
C &= \sum_{i\ne j}\left[p_{ij}\log\left(\frac{p_{ij}}{q_{ij}}\right)+(1-p_{ij})\log\left(\frac{1-p_{ij}}{1-q_{ij}}\right)\right].
\end{align}
\revise{%
Ignoring terms that are constant with respect to the embedding yields attractive and repulsive forces. The attraction term is
}
\begin{align}\label{Eq:Attraction}
a_{ij}p_{ij}=\frac{-2ab\lVert\vy_i-\vy_j\rVert^{2(b-1)}}{1+a\lVert\vy_i-\vy_j\rVert^{2b}}p_{ij},
\end{align}
\revise{%
and the repulsion term is
}
\begin{align}\label{Eq:Repulsion}
r_{ij}(1-p_{ij})=\frac{b}{(\epsilon+\lVert\vy_i-\vy_j\rVert^2)(1+a\lVert\vy_i-\vy_j\rVert^{2b})}(1-p_{ij}).
\end{align}
\revise{%
This graph-based formulation makes UMAP suitable for our extension: instead of concatenating all descriptors into one distance, we can build a fuzzy graph per attribute and combine the resulting membership strengths in the objective.
}

\paragraph*{\revise{Multi-Graph and Guided Embeddings.}}
\revise{%
MultiMAP~\cite{Jain2021} extends UMAP by merging graphs from multiple datasets into a shared fuzzy simplicial set. Our setting differs because the same watches are described by multiple heterogeneous attributes. We keep the attribute graphs separate until the objective stage, which allows explicit control over their relative influence. LAMP~\cite{Joia2011} provides interactive control points for steering projections; our heptagonal-type layout plays a similar guiding role, but it is derived from class labels and decays during optimization to preserve local neighborhoods. The method is therefore positioned between fully automatic projection and user-steered layout: type labels guide global orientation, while attribute-specific graphs determine local similarity.
}

\begin{figure*}[t]
\centering
\begin{tikzpicture}

\node[anchor=north west] (img) at (0,0)
{\includegraphics[width=0.98\linewidth]{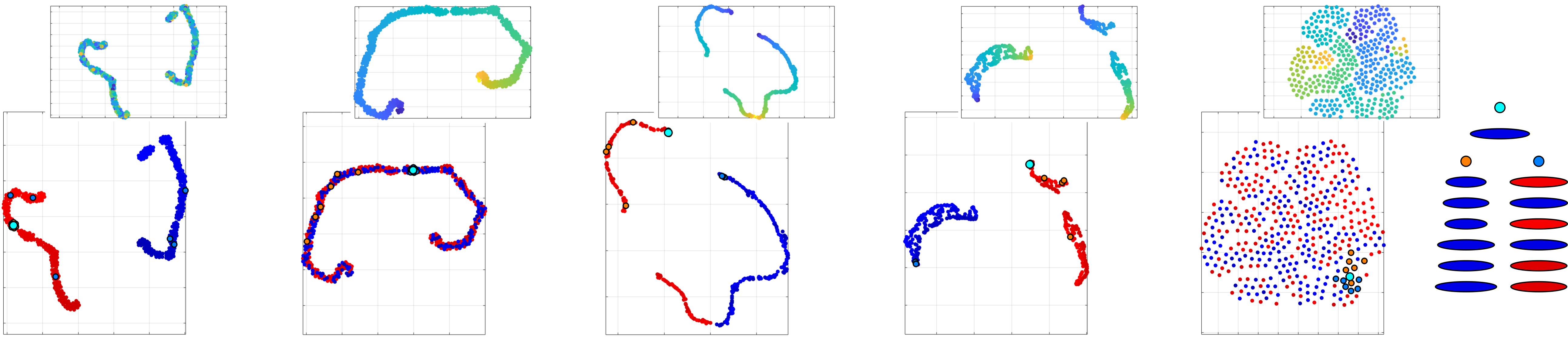}};

\begin{scope}[
x={($(img.south east)-(img.south west)$)},
y={($(img.south west)-(img.north west)$)}
]
\node at (0.02,0.1) {(a)};
\node at (0.2,0.1) {(b)};
\node at (0.4,0.1) {(c)};
\node at (0.59,0.1) {(d)};
\node at (0.775,0.1) {(e)};
\node at (0.92,0.9) {nearest };
\node at (0.92,0.96) {color};
\node at (0.98,0.9) {nearest };
\node at (0.98,0.96) {shape};
\end{scope}

\end{tikzpicture}
\caption{Motivational example for multi-attribute embedding. Each column (a--e) shows high-dimensional attributes (top: shape; bottom: RGB color) and their 2D embedding. Orange and blue markers denote color and shape neighbors of the cyan point. (a--b) Traditional UMAP using a single attribute preserves only color or shape. (c--d) Traditional UMAP with stacked, equally-weighted attributes, with and without normalization, yields plausible clusters but incorrect neighborhoods. (e) Our multi-attribute UMAP preserves both color and shape neighbors. See the supplemental material for more details.}%
\label{fig:motivationalExample}
\vspace{-6px}
\end{figure*}
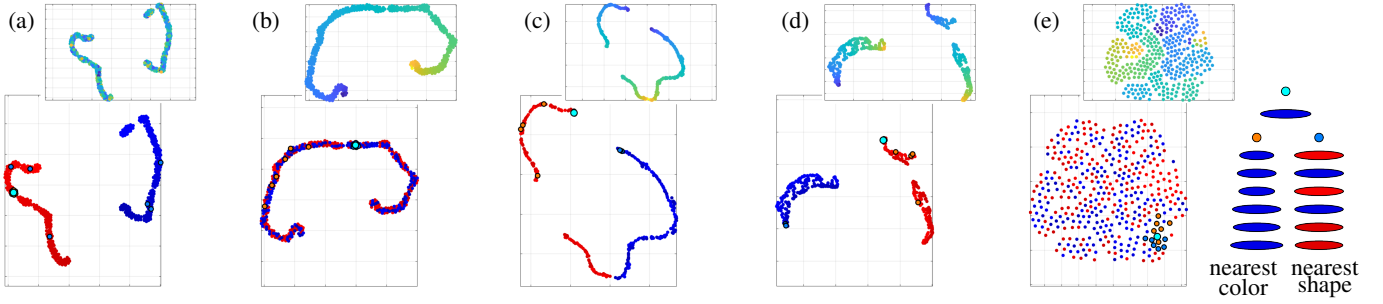

\newpage

\revise{%
\section{A Simple Example}
\label{sec:simple-example}

To motivate the problem addressed in this paper, we consider a simple synthetic example consisting of ellipses with two heterogeneous attributes: color and shape. Each ellipse is described by an RGB color vector and by a scalar elongation factor controlling its geometry. The colors are sampled from two clearly separated RGB groups, representing warm and cold colors, while the elongation values are sampled independently from a Gaussian distribution. Thus, color and shape define two different neighborhood structures over the same set of objects. A pair of ellipses can be similar in color but dissimilar in shape, or similar in shape but dissimilar in color.

\autoref{fig:motivationalExample} illustrates the resulting challenge. When standard UMAP is applied to only one attribute, the corresponding structure is preserved, but the other attribute is ignored: the color-only embedding preserves chromatic neighborhoods, whereas the shape-only embedding preserves neighborhoods induced by elongation. A common alternative is to concatenate both attributes into a single feature vector and apply UMAP to the joint representation. However, this treats all feature dimensions as part of one homogeneous metric space. As shown in \autoref{fig:motivationalExample}(c--d), this can yield visually plausible clusters, but the local neighborhoods around a query object are not necessarily correct. Even after standard score normalization, the embedding still depends on how the heterogeneous attributes interact under a single distance metric.

Our approach avoids this premature fusion. Instead of concatenating color and shape into one feature vector, we construct separate neighborhood graphs for the individual attributes and combine these graphs using explicit user-defined weights. In the example, the orange and blue markers show the color and shape neighbors of the cyan query point, respectively. The desired embedding should preserve representatives from both neighborhoods according to the selected attribute weights. As shown in \autoref{fig:motivationalExample}(e), our multi-attribute UMAP better preserves both neighborhood types in the final 2D layout. The supplemental material provides the full construction of the synthetic data, the weighting scheme, and the quantitative neighborhood-preservation analysis.
}

\section{Multi-Attribute Latent Space for Visual Analysis}
\label{Sec:DesignRationale}

\revise{%
This section connects the domain requirements to the embedding and system design. The central design decision is to distinguish between attributes that should structure the global overview and attributes that should define local neighborhoods. Watch type is a semantic and functional category; users often start with it when deciding whether a model is a diver, chronograph, dress watch, or another type. Color and dial design are visual attributes; they are more useful for comparing alternatives within or across types. A single feature space cannot represent these roles transparently, and a pure grid or faceted layout would not reveal continuous visual neighborhoods.
}

\revise{%
We therefore make three design commitments. First, the latent space should expose global type regions without implying that all distances between types have precise semantic meaning. Second, local neighborhoods should be computed separately for color and design, because these attributes can disagree and users may want to emphasize one over the other. Third, the map should be coupled with interaction: metadata filters, detail views, and search-by-example are necessary because a 2D projection is an abstraction and not a complete model of product relevance.
}

\revise{%
To instantiate this design, we proceed in five steps. We introduce domain terminology (\autoref{sec:DomainAnalysis}), derive requirements from interviews and online observations (\autoref{Sec:Requirement}), extract high-dimensional attributes (\autoref{Sec:FeatureExtraction}), define the multi-attribute UMAP objective (\autoref{Sec:LatentSpace}), and describe the interactive visualization (\autoref{sec:Visualization}).
}

\subsection{Domain Terminology}
\label{sec:DomainAnalysis}

\begin{figure}[t]%
    \centering%
    \includegraphics[width=\linewidth]{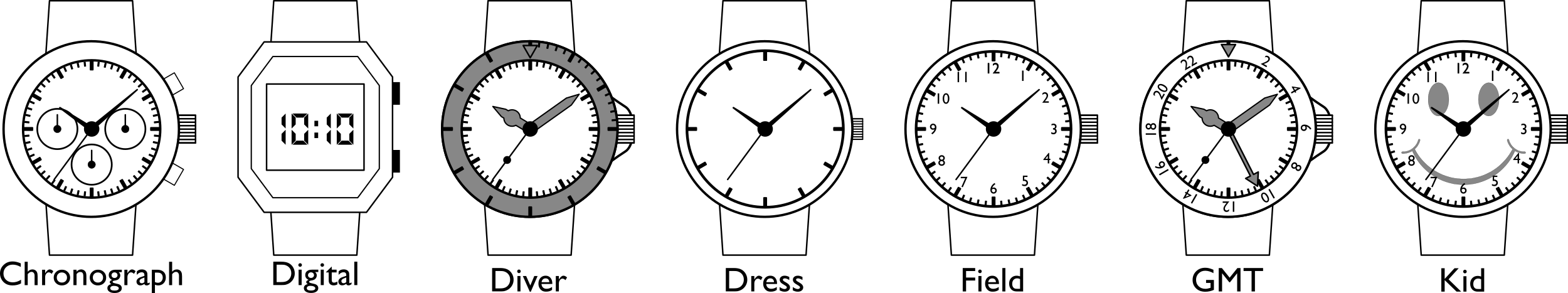}%
    \caption{Schematic illustration of the considered watch types.}%
    \label{fig:watch-characterization}%
\end{figure}%

For our study, we use the following practical taxonomy of watches. 
An illustration for each type is provided in \autoref{fig:watch-characterization}.
\begin{itemize}[topsep=2pt] 
\item[1.] \textbf{Chronograph}:
A chronograph is a multifunctional watch with stopwatch controls and sub-dials for time intervals, often featuring a tachymeter bezel and linked to motorsports and aviation.

\item[2.] \textbf{Digital Watch}:
A digital watch utilizes LCD/LED displays to show time and features alarms, timers, etc. They are lightweight, durable, and valued for practicality and affordability.

\item[3.] \textbf{Diver Watch}:
Diver watches were originally developed for underwater exploration. They typically provide high water resistance ($\ge 200\,m$), a unidirectional bezel for timing, luminous markers for visibility, and robust cases with screw-down crowns and seals.

\item[4.] \textbf{Dress Watch}:
Dress watches are formal timepieces with minimalist designs, clean dials, slim profiles, and refined materials. They often feature minimal complications (i.e., dial details), emphasizing elegance and discretion.

\item[5.] \textbf{Field/Pilot Watch}:
Field and pilot watches originate from military use and prioritize durability, legibility, and simplicity, featuring large Arabic numerals (many pair a 12-hour scale with a 13-24-hour inner ring), compact cases, and oversized crowns. 
\item[6.] \textbf{GMT Watch}:
\revise{%
GMT watches enable tracking of multiple time zones through an additional 24-hour hand and a rotatable bezel. 
Originally developed for pilots and frequent travelers, they provide a practical solution for coordinating time across regions.%
}%
\item[7.] \textbf{Kids Watch}:
\revise{%
A kids' watch prioritizes ease of use, clear readability, and playful design. 
They often include educational or themed elements such as labeled hands or cartoon motifs.
}
\end{itemize}

\begin{figure*}[t]
    \centering
    \includegraphics[width=0.99\linewidth]{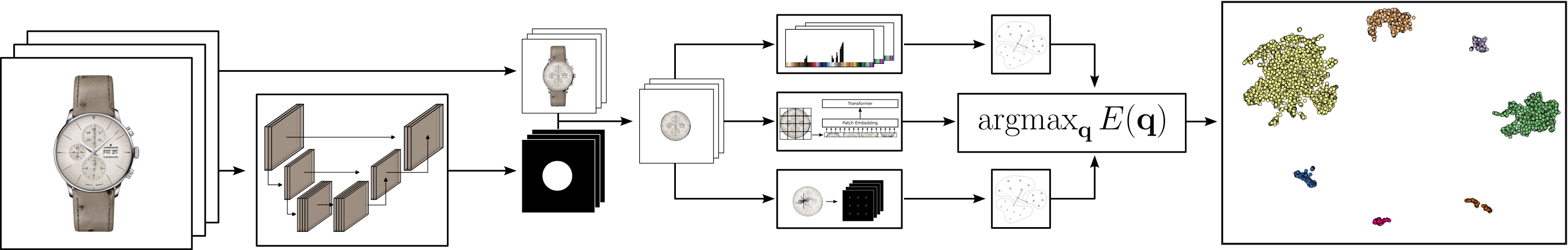}
    \caption{    Attribute extraction pipeline. For each watch image, we first segment the dial with a U-Net. Afterwards, a ViT is used to determine the watch type. Finally, a color attribute vector and a design attribute vector are constructed using color palette projection and HOG.
    }
    \label{fig:pipeline}
\end{figure*}

\newpage
\subsection{Requirement Analysis}
\label{Sec:Requirement}

\revise{%
We conducted a requirement analysis to understand how users compare watches and which visualization abstractions are appropriate. The analysis combined (1) inspection of watch retail platforms and enthusiast communities, where comparison is commonly based on brand pages, faceted filters, ranked lists, and side-by-side images, and (2) four structured interviews with two experts and two novices. Experts had several years of experience in collecting, reviewing, selling, or designing watches; novices had a general interest in consumer products but limited watch-specific experience. Each interview lasted 30--45 minutes and focused on search strategies, comparison criteria, failure cases of existing interfaces, and desired forms of visual exploration.
}

\revise{%
The interviews showed that watch exploration is rarely a single-step lookup. Users first narrow the space by intended use or type, then compare visual alternatives, and finally apply practical constraints. Experts described workflows such as starting from a known reference model, inspecting similar dial layouts or complications, and looking for alternatives across brands. Novices described broader strategies, for example, starting with color or a visual impression and then learning which watch types match that preference. Existing catalog interfaces support the final filtering step, but they do not expose visual neighborhoods or make it easy to move from one visually similar alternative to another.
}%
\revise{%
We used a semi-structured interview format to elicit these workflows and pain points. The following prompts guided the interviews, but were not intended as separate evaluation criteria:
}
\begin{itemize}[noitemsep, topsep=2pt]
    \item \revise{\textit{How do you usually search for or compare watches online?}}
    \item \revise{\textit{Which visual attributes make two watches feel similar/different?}}
    \item \revise{\textit{When do existing filters, grids, or lists fail to support comparison?}}
    \item \revise{\textit{Would you start from a type, a color, a reference image, or a practical constraint?}}
    \item \revise{\textit{What information would you need to trust a visual recommendation or neighborhood?}}
\end{itemize}
\revise{%
Based on the responses in the interviews, we derived the following three requirements. 
}

\vspace{1ex}\noindent%
\customlabel{W:type}{\textbf{W1}} \textbf{Global type overview.}
\revise{%
The visualization should make broad watch types visible because type captures function, user expectations, and design conventions. This later motivates our addition of a class-aware layout term and the membership-neighborhood graph. The global layout serves as an overview rather than as a precise type-to-type distance representation.
}
\vspace{1ex}\\
\customlabel{W:color}{\textbf{W2}} \textbf{Color-based local similarity.}
\revise{%
Color strongly shapes first impressions and is a common entry point for novice users. Watches with similar dial color palettes should be easy to find, regardless of brand or watch type. This motivates introducing a separate color descriptor and a color-specific neighborhood graph, rather than treating color solely as metadata.
}
\vspace{1ex}\\
\customlabel{W:design}{\textbf{W3}} \textbf{Design-based local similarity.}
\revise{%
Dial layout, hands, indices, complications, and texture define a deeper level of stylistic similarity. These cues are important for expert comparison and for search-by-example. Analogous to color-based similarity, this calls for a separate design descriptor and a design-specific neighborhood graph.
}
 \vspace{1ex}\\
 \customlabel{W:hybrid}{\textbf{W4}} \textbf{Hybrid exploration with constraints.}
 \revise{%
 Visual similarity is only one part of product exploration. Users also need to consider practical constraints such as price, brand, case size, and other metadata. The latent map should therefore be coordinated with conventional filtering and detail inspection, allowing users to move from visual discovery to more targeted product comparison.}

\revise{%
These requirements also clarify the role of the latent-space visualization. The map is most appropriate for discovery, comparison, and query refinement. It is not intended to replace precise filtering, ranked retrieval, or expert judgment about mechanical quality. The following section describes the attribute extraction pipeline used to instantiate the requirements \autoref{W:type}--\autoref{W:design}.
}

\section{High-Dimensional Attribute Extraction} 
\label{Sec:FeatureExtraction}

\revise{ 
To construct the visual and semantic representation used by the latent-space visualization, our system takes a large collection of watch images as input. The requirement analysis in \autoref{Sec:Requirement} identified three characteristics that should be represented explicitly in the embedding: watch type (\autoref{W:type}), watch color (\autoref{W:color}), and watch design (\autoref{W:design}). These requirements describe the visual and semantic structure of the latent space. The fourth requirement, hybrid exploration with practical constraints (\autoref{W:hybrid}), concerns the coordination of the latent map with metadata filters and detail inspection and is therefore addressed by the visualization system in \autoref{sec:Visualization}.} 
\revise{%
Since watch type, dial color, and dial design are not directly available in a consistent form from the raw images, we apply learning-based classification and attribute extraction to derive them automatically. The attribute extraction pipeline is illustrated in~\autoref{fig:pipeline}. 
}
First, the data is prepared and cleaned (\autoref{Sec:Acquisition}). Next, the dial is segmented with a U-Net (\autoref{Sec:Segmentation}), and a ViT classifies the watch type (\autoref{Sec:Category}). Finally, color attributes are derived from a reference palette (\autoref{Sec:ColorFeatures}) and design attributes are computed using HOG (\autoref{Sec:DesignFeatures}).

\subsection{Data Acquisition \& Duplicate Removal} 
\label{Sec:Acquisition}
We collected a diverse set of watch images from Chrono24~\cite{chrono24}, where sellers upload their own photographs, resulting in varied viewing angles, lighting conditions, and backgrounds. In total, we downloaded over 33,000 labeled images covering multiple watch types.
We removed duplicates using a combination of cryptographic and perceptual hashing. Each image was converted to an MD5 hash to detect bitwise identical copies. Additionally, perceptual hashing (pHash) was applied by converting images to grayscale, downsampling them to $8\times8$, and comparing their mean intensities to generate a binary descriptor that captures coarse luminance patterns.
Images with matching perceptual hashes were treated as duplicates, resulting in 32,861 labeled images for training and evaluation. The dataset shows a natural class imbalance: chronographs (9,535), diver watches (9,383), and GMT watches (7,700) are most common, while field (1,194), dress (2,087), digital (1,435), and kids watches (1,527) are less represented.
\definecolor{bblue}{HTML}{4F81BD}
\definecolor{ggreen}{HTML}{9BBB59}
\definecolor{ppurple}{HTML}{9F4C7C}

\definecolor{oorange}{HTML}{dd5400}
\definecolor{yyellow}{HTML}{edb120}
\definecolor{lbblue}{HTML}{2fbeef}

\subsection{Dial Segmentation}
\label{Sec:Segmentation}
\revise{%
We trained a U-Net~\cite{Ronneberger2015} for dial segmentation on 1,263 manually annotated images. The subset was sampled to cover the seven watch types and to include different dial colors, case shapes, lighting conditions, and image compositions. We use the term \emph{dial} to refer only to the visible watch face inside the case. Wristbands or bracelet elements are not segmented and are not treated as a separate feature class. This is a deliberate design choice because watch straps can typically be exchanged independently of the watch head and therefore do not provide a stable descriptor of the watch model itself. The segmentation step is thus used to restrict subsequent color and design descriptors to the dial region, while wristband appearance is excluded from the feature computation.
}

\revise{%
The dial masks, including the bezel, were annotated manually by drawing the outer visible boundary of the dial for each image. Internal dial elements such as hands, indices, numerals, subdials, logos, and date windows were included as part of the dial region because they contribute to the visual design of the watch face. Since the goal was a binary foreground mask for the dial rather than a multi-class semantic segmentation, no additional labels were assigned to hands, markers, subdials, or wristband components.
}

\revise{%
The initial dial masks were annotated by the main author. For quality assurance, all masks were subsequently inspected by a co-author as overlays on the original images. Cases with visible boundary errors, missing dial regions, accidental inclusion of case or wristband pixels, or ambiguous dial boundaries were discussed between the authors and corrected when necessary. This review-and-discussion process was used to ensure a consistent interpretation of the dial boundary across the dataset. 
}

\revise{%
Only the luminance channel was used for segmentation because the boundary between dial and case is mainly a shape and contrast cue; color is introduced later in the dedicated color descriptor. The annotated images were split into 80\,\% training, 10\,\% validation, and 10\,\% test sets. Data augmentation was applied online during training to avoid a fixed enlarged dataset: each epoch sampled random affine transformations for the image-mask pair, including shifts, scaling in $[0.5,1.5]$, and rotations between $-90^\circ$ and $90^\circ$. The augmented images were resized to $64\times64$ pixels.
}

\revise{%
We evaluated U-Net variants with different encoder depths using accuracy and intersection over union (IoU) on the held-out test set. The goal was not only to maximize segmentation accuracy, but also to keep the segmentation stage fast enough for search-by-example insertion. \autoref{Tab:unet-accuracy} summarizes the results. Depths three and four achieved similar IoU, so the shallower option is preferable when runtime is prioritized.
}

\setlength{\tabcolsep}{5pt}
\begin{table}[h]
\centering
\caption{Comparison of U-Net encoder depths. Accuracy and intersection over union (IoU) are reported for dial and background segmentation.}
\label{Tab:unet-accuracy}
\begin{tabular}{ l|cc|cc|cc  }
\toprule
 \multicolumn{1}{l|}{Depth ($d$)} & \multicolumn{2}{c|}{$d=2$} & \multicolumn{2}{c|}{$d=3$} & \multicolumn{2}{c}{$d=4$} \\ 
 Metrics    & Accur.    & IoU       & Accur.    & IoU       & Accur.    & IoU \\       
 \midrule
 Dial       & 0.94      & 0.88      & 0.95      & 0.91      & 0.95      & 0.91 \\     
 Backgr.    & 0.99      & 0.97      & 0.99      & 0.98      & 0.99      & 0.98 \\ \bottomrule

\end{tabular}
 \vspace{-7px}
\end{table}

\subsection{Identification of Watch Type}
\label{Sec:Category}
\revise{%
To classify user-provided images (\autoref{W:type}), we fine-tune pretrained ViT models~\cite{Dosovitskiy2021}. ViTs process images as sequences of $16\times16$ patches and apply self-attention blocks~\cite{vaswani2017}, enabling global context modeling. The base model was initialized with ImageNet-21k weights~\cite{Ridnik2021}.
}%
For our task, we replaced the classification head with a fully connected layer matching the seven watch-type labels. The original transformer layers were retained and optionally frozen during early training to preserve learned representations, and then unfrozen for fine-tuning. As in the U-Net training (\autoref{Sec:Segmentation}), data augmentation included random shifting, scaling, and rotation. We used a stratified split for training (80\,\%), validation (10\,\%), and test (10\,\%) sets~\cite{sechidis2011}, resizing images to the ViT input resolution (e.g., $384\times384$).
Training used the Adam optimizer with early stopping based on the validation loss. Performance was evaluated using match accuracy and per-label precision on the test set. We compared three architectures of increasing size: Tiny-16 (22.2 MB), Small-16 (84.7 MB), and Base-16 (331.4 MB), see~\autoref{Tab:Category}.

\begin{table}[ht]
\centering
\caption{Evaluation of accuracy for the three ViT architectures. Below, the average per-sample accuracy (overall) is listed.}
\label{Tab:Category}
\begin{tabular}{ l|ccc  }
 \toprule
 Category   & tiny-16   & small-16  & base-16 \\ \midrule
 Chronograph& 92.66\,\%   & 92.86\,\%   & 94.33\,\% \\
 Digital    & 83.92\,\%   & 76.22\,\%   & 85.32\,\%\\
 Diver      & 86.99\,\%   & 91.68\,\%   & 92.54\,\%\\
 Dress      & 83.73\,\%   & 78.95\,\%   & 83.25\,\%\\
 Field      & 82.35\,\%   & 82.35\,\%   & 83.19\,\%\\
 GMT        & 85.58\,\%   & 88.96\,\%   & 84.94\,\%\\
 Kids       & 88.98\,\%   & 88.24\,\%   & 85.62\,\%\\ 
 \midrule
 Overall    & 87.88\,\%   & 89.41\,\%   & 89.71\,\%\\ \bottomrule
\end{tabular}
 \vspace{-7px}
\end{table}

\subsection{Color-Based Attribute Representation}
\label{Sec:ColorFeatures}
We construct a discriminative attribute vector capturing the dominant color composition of each image while remaining comparable across images~\autoref{W:color}. For this, we apply a two-stage color abstraction approach: first, extracting image-specific dominant colors through clustering, and then deriving a global reference color palette shared across the data.

\paragraph*{Image-Level Dominant Color Extraction.}
First, we extract the dominant colors of each image. Due to its approximate linearity, images are transformed from RGB to the perceptually uniform CIELAB color space. We then apply $k$-means clustering to obtain a compact set of representative colors. Empirically, $k=8$ colors are sufficient to describe the perceptual color composition of an image without overfitting to minor variations. The resulting centroids \( \{c^i_{1}, c^i_2, \dots, c^i_8\} \in \R^3 \) represent the dominant colors of the $i$th image.

\paragraph*{Global Reference Palette Generation.}
To facilitate consistent attribute extraction across the dataset, we construct a global reference palette that consolidates dominant colors from all images. 
Specifically, we aggregate the set of image-level centroids across the entire dataset, resulting in a pool of candidate colors $\mathcal{C}$: 
{\setlength{\abovedisplayskip}{4pt}
\setlength{\abovedisplayshortskip}{4pt}
\setlength{\belowdisplayskip}{4pt}
\setlength{\belowdisplayshortskip}{4pt}
\[
\mathcal{C} = \bigcup_{j=1}^N\{c^i_{1}, c^i_2, \dots, c^i_8\}.
\]}

\begin{figure}[t]
    \centering
    \includegraphics[width=0.95\linewidth]{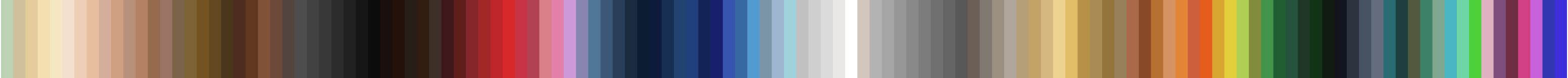}
    \caption{The \( k=128 \) colors extracted from the watch images. }
    \label{fig:Palette_trim}
\end{figure}

This set contains \(8N\) elements for \(N\) images. We again apply \(k\)-means clustering with \(k=128\) over \(\mathcal{C}\) to obtain 128 \emph{globally} representative color centroids 
\(\mathcal{R}=\{r_1,\dots,r_{128}\}\), shown in~\autoref{fig:Palette_trim}. These centroids define a reference color vocabulary representing common regions of the perceptual color space. The parameter \(k=128\) was chosen empirically to balance expressiveness and compactness.

\paragraph*{Color Attribute Vector Construction.}

\begin{figure}[t]
    \centering
    \includegraphics[width=0.8\linewidth]{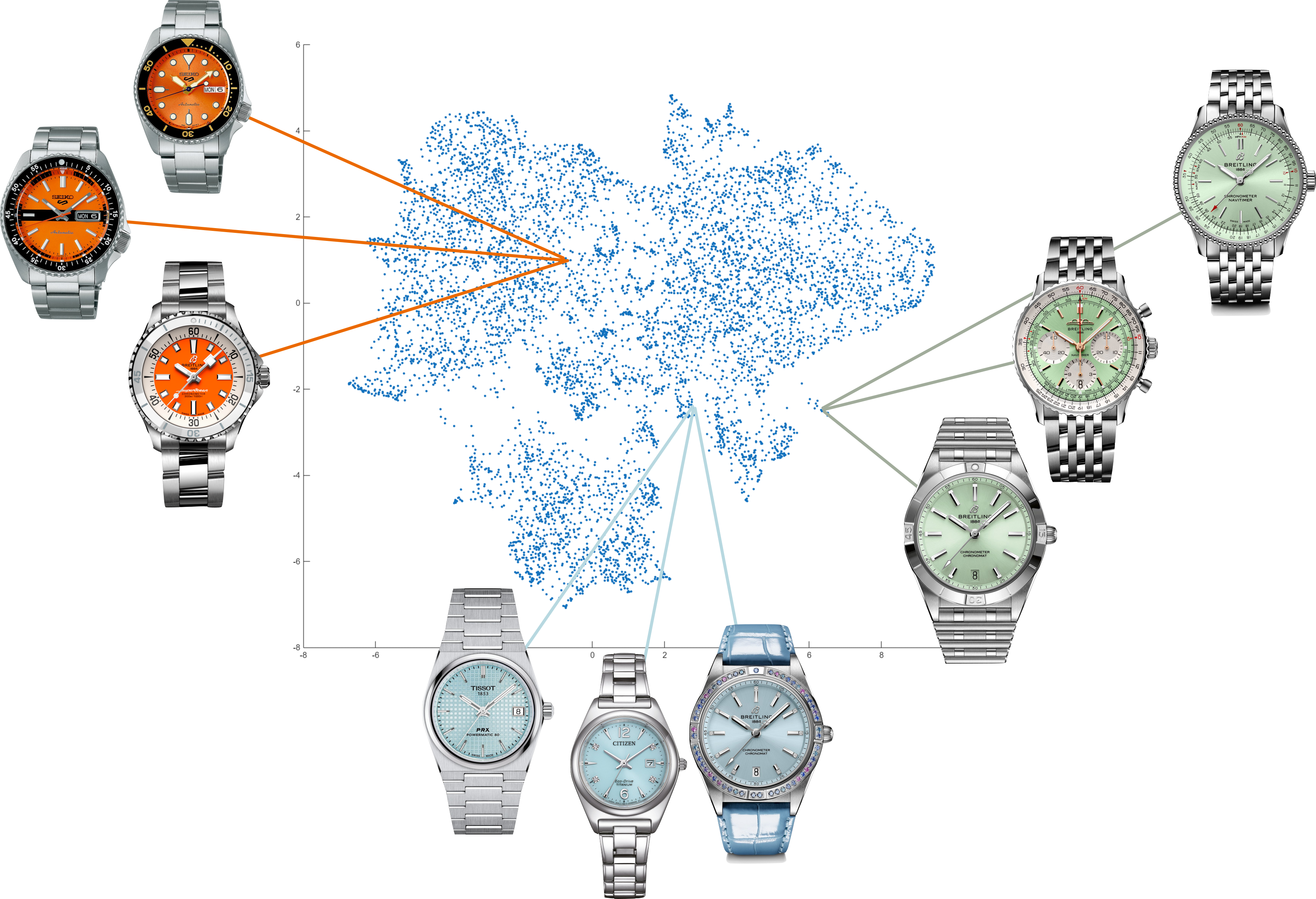}
    \caption{After determining a \emph{color} attribute vector for every image, UMAP was applied to embed the images in a 2D space. }
    \label{fig:LatentSpace_Color}
\end{figure}

For each image, we compute a fixed-length color attribute vector by projecting LAB pixel colors onto the global reference palette \( \mathcal{R} \). Each pixel is assigned to its nearest reference color \( r_i \) using Euclidean distance in CIELAB space, and the assignment histogram is normalized to sum to one. The resulting 128-dimensional vector \( \mathbf{f} \in \R^{128} \) encodes the proportion of pixels \( f_i \) associated with each reference color \( r_i \). To demonstrate that this descriptor is consistent and interpretable, we applied a standard UMAP~\cite{McInnes2020} to the color attribute vectors of all images, see~\autoref{fig:LatentSpace_Color}.

\begin{figure}[t]
    \centering
    \includegraphics[width=0.8\linewidth]{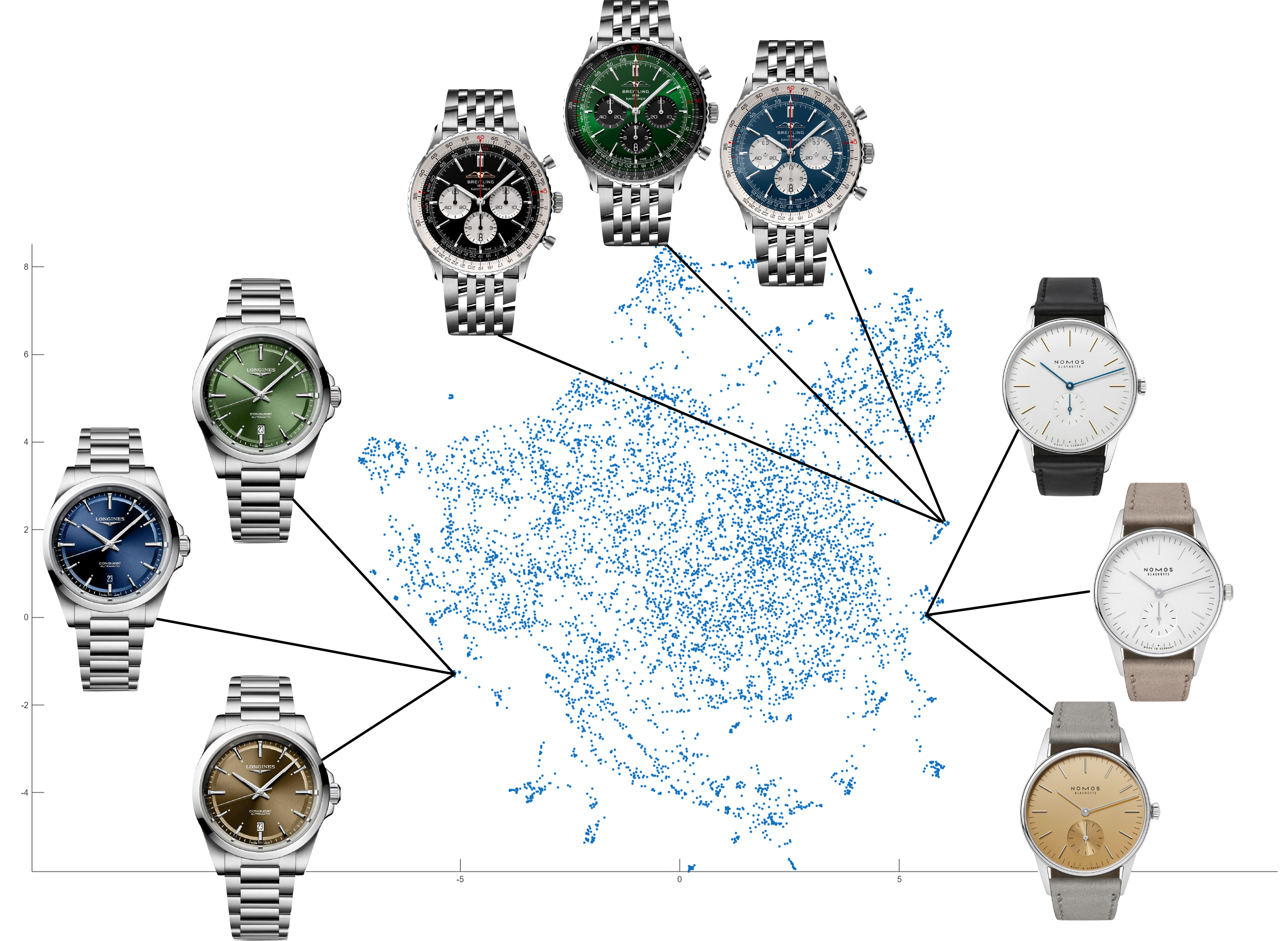}
    \caption{After determining a \emph{design} attribute vector for every image, UMAP was applied to embed the images in a 2D space.}
    \label{fig:LatentSpace_HOG}
\end{figure}

\subsection{Design-based Attribute Representation}
\label{Sec:DesignFeatures}

\revise{%
To form an attribute vector for dial design (\autoref{W:design}), we use HOG attributes~\cite{Dalal2005,Liu2014}. This choice is deliberately conservative. Deep features could encode more semantic information, but they may also entangle brand, photography style, or type labels with the design cues we want to inspect. HOG instead emphasizes local gradient structure and is therefore suitable for cues such as subdials, hands, indices, numerals, and textured patterns. It does not capture every aspect of design, such as material finish or brand-specific case geometry, but it provides an interpretable descriptor that complements the color histogram.
}

\revise{%
Each watch image is resized to $384\times384$ pixels and converted to grayscale after dial segmentation (cf.~\autoref{Sec:Segmentation}). We use a cell size of $96\times96$ pixels, yielding a $4\times4$ grid. With nine orientation bins and a block size of $2\times2$, the resulting descriptor has length $324$. Smaller cells increased descriptor length and made neighborhoods sensitive to small alignment errors; larger cells oversmoothed subdial and marker structure. \autoref{fig:LatentSpace_HOG} shows a standard UMAP projection of the HOG descriptors, illustrating that design neighborhoods differ from color neighborhoods.
}

\section{Multi-Attribute UMAP}
\label{Sec:LatentSpace}

\paragraph{\revise{Choice of Underlying Embedding Model.}}
\revise{We use UMAP as the basis for our embedding because its graph-based formulation directly matches the requirements of our multi-attribute setting. In contrast to projection methods that operate primarily on a single distance matrix or optimize a fixed low-dimensional stress objective, UMAP explicitly separates the construction of a high-dimensional fuzzy neighborhood graph from the optimization of the low-dimensional embedding. This separation allows us to replace the single neighborhood graph with multiple attribute-specific graphs, each encoding a different notion of similarity, and to combine their membership strengths in a controlled manner. The resulting formulation preserves the standard UMAP objective while making the contribution of color, design, and type-aware neighborhoods explicit.
Other dimensionality-reduction methods could serve as visualization baselines, but they are less suitable as the basis for our proposed extension. PCA is linear and therefore insufficient for the non-linear visual neighborhoods considered here. t-SNE is effective for local cluster visualization, but displays less stable global structure needed for layout-guided exploration. In \autoref{Sec:Conclusion}, we provide preliminary results for a t-SNE variant of our multi-attribute embedding. LAMP and related user-steerable projections support interactive refinement, but require control points and are not primarily designed to integrate multiple automatically extracted attribute graphs. We therefore treat these methods as relevant baselines rather than as equally suitable foundations for the proposed method.
Our goal is not to claim that UMAP is universally superior to other dimensionality reduction methods. Instead, we use UMAP because its fuzzy simplicial set representation provides a natural insertion point for multi-attribute neighborhood modeling and class-aware layout constraints. The evaluation, therefore, compares our UMAP-based formulation against standard UMAP applications and other dimensionality-reduction baselines to quantify the trade-offs introduced by the proposed modifications.}

\paragraph{\revise{Approach to Multi-Attribute Embedding.}}
\revise{%
In \autoref{Sec:ColorFeatures} and \autoref{Sec:DesignFeatures}, we derived separate descriptors for color and dial design. In \autoref{Sec:Category}, we predicted the watch type used for global organization. Applying standard UMAP to each descriptor produces different maps (\autoref{fig:LatentSpace_Color} and \autoref{fig:LatentSpace_HOG}), which is expected because each descriptor defines a different notion of similarity. Concatenating the descriptors would obscure this distinction and could distort neighborhoods when a single descriptor dominates the distance metric.
}%
\revise{%
Our extension keeps the descriptors separate until the graph-combination stage. For every attribute $f\in\cF$, we build a fuzzy neighborhood graph and combine its edge weights with a user-controlled weight $\gamma_f$. We then add two type-aware mechanisms: membership-constrained neighborhoods and a decaying global layout energy. The resulting optimization supports three use cases: color-driven, design-driven, and balanced exploration. The parameters are intentionally exposed because there is no single correct definition of similarity for product exploration.
}

\subsection{Multiple Attribute Vectors in UMAP}
\label{sec:MultipleFeatureVectorsUMAP}
Following the UMAP formulation, each attribute induces its own neighborhood distribution in the high-dimensional space (cf.~\autoref{Eq:MembershipProbability}). For each pair of watches $(i,j)$ and attribute $f\in\cF$, we compute a membership strength $p^f_{ij}$ that expresses how strongly watch $j$ belongs to the neighborhood of watch $i$ w.r.t. attribute $f$. This yields one fuzzy neighborhood graph per attribute, each encoding a distinct notion of similarity. Our goal is to find a common 2D embedding whose low-dimensional similarities $q_{ij}$ are simultaneously compatible with these attribute-specific graphs. To this end, we insert the weighted memberships $\gamma_f p^f_{ij}$, with $\gamma_f\in[0,1]$, into the UMAP cross-entropy objective. The resulting formulation exposes the following decomposition:
{\setlength{\abovedisplayskip}{3pt}
\setlength{\abovedisplayshortskip}{3pt}
\setlength{\belowdisplayskip}{3pt}
\setlength{\belowdisplayshortskip}{3pt}
\begin{align}\label{Eq:CrossEntropy_Mixed}
    C &= \sum_{f\in \cF}\gamma_f\sum_{i \neq j} \Big[p^f_{ij} \log\left(\frac{p^f_{ij}}{q_{ij}}\right)  + (1 - p^f_{ij}) \log\left(\frac{1-p^f_{ij}}{1-q_{ij}}\right) \Big]\nonumber\\
    &= \sum_{f\in \cF}\gamma_f\sum_{i \neq j} \Big[ \underbrace{p^f_{ij} \log (p^f_{ij})  + (1 - p^f_{ij}) \log (1- p^f_{ij})}_{C^f_{ij}} -\nonumber\\ 
&\qquad\left(p^f_{ij} \log (q_{ij})  + (1 - p^f_{ij}) \log (1-q_{ij})\right)  \Big]\nonumber\\ 
&= \sum_{f\in \cF}\gamma_f\sum_{i \neq j} C^f_{ij} - \nonumber \\
&\qquad\sum_{i \neq j}\Big[ \Big(\sum_{f\in \cF} \gamma_f p^f_{ij}\Big) \log (q_{ij})  +\Big(\sum_{f\in \cF} \gamma_f - \gamma_f p^f_{ij}\Big) \log (1-q_{ij})\Big].
\end{align}}%
After rearranging as done above, we can define an overall probability $P_{ij}$ in the high-dimensional graph:
{\setlength{\abovedisplayskip}{4pt}
\setlength{\abovedisplayshortskip}{4pt}
\setlength{\belowdisplayskip}{4pt}
\setlength{\belowdisplayshortskip}{4pt}
\begin{align}
\label{Eq:ProbabilityGraph}
    P_{ij}=\sum_{f\in\cF}\gamma_fp^f_{ij},
\end{align}}%
which simplifies~\autoref{Eq:CrossEntropy_Mixed} greatly:
{\setlength{\abovedisplayskip}{4pt}
\setlength{\abovedisplayshortskip}{4pt}
\setlength{\belowdisplayskip}{4pt}
\setlength{\belowdisplayshortskip}{4pt}
\begin{align}\label{Eq:CrossEntropy_Mixed_Final}
    C &= \sum_{f\in \cF}\sum_{i \neq j} \gamma_f C^f_{ij} - \sum_{i \neq j}\Big[ P_{ij} \log (q_{ij})  +( 1 - P_{ij}) \log (1-q_{ij})\Big].
\end{align}}%
Thus, the resulting edge weight in the high-dimensional graph is a linear combination of attribute vectors.
To find a low-dimensional embedding, we minimize~\autoref{Eq:CrossEntropy_Mixed_Final}, or, equivalently, we maximize:
{\setlength{\abovedisplayskip}{4pt}
\setlength{\abovedisplayshortskip}{4pt}
\setlength{\belowdisplayskip}{4pt}
\setlength{\belowdisplayshortskip}{4pt}
\begin{align}
\label{Eq:CrossEntropy_Simple}
    E_{\textrm{entropy}}
    &= \sum_{i \neq j}\Big[ P_{ij} \log (q_{ij})  +( 1 - P_{ij}) \log (1-q_{ij})\Big].
\end{align}}

\subsection{Class-Aware Embedding}
\label{sec:ClusterAwareEmbedding}
Standard UMAP constructs a graph based solely on nearest neighbors, independent of any underlying class or category structure.  

\begin{figure}[t]
    \centering
    \includegraphics[width=0.45\linewidth]{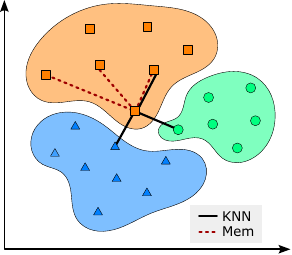}\hfill
    \raisebox{0.2cm}{\includegraphics[width=0.4\linewidth]{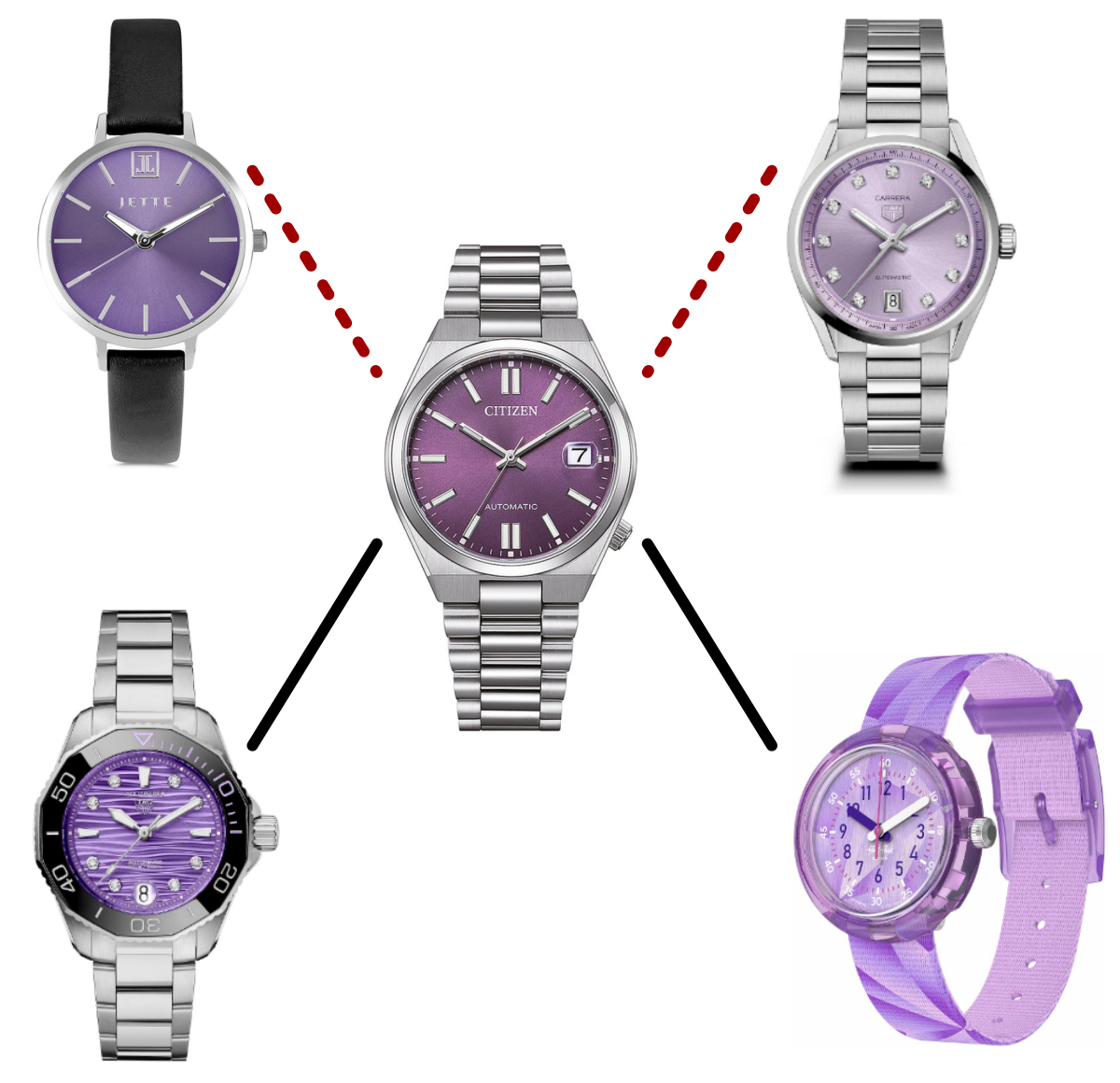}}
    \caption{Left: Classical nearest neighbors and class-based neighbors for each high-dimensional point. Right: A purple dress watch (center) with two class neighbors (dress watches, top) and two regular KNN neighbors (diver and kids watch, bottom).}
    \label{fig:KNN_MEM}
       
\end{figure}

\paragraph*{Membership Neighbors.}
To incorporate class labels, we additionally build a graph based on the $k$-nearest \emph{membership neighbors}, considering only watches with the same type as neighbors, cf. \autoref{Sec:Category}.
This modification allows us to update \autoref{Eq:ProbabilityGraph} as follows:
{\setlength{\abovedisplayskip}{4pt}
\setlength{\abovedisplayshortskip}{4pt}
\setlength{\belowdisplayskip}{4pt}
\setlength{\belowdisplayshortskip}{4pt}
\begin{align}
\label{Eq:ProbabilityGraph-updated}
    P_{ij} = \sum_{f \in \mathcal{F}} \sum_{g \in \{\text{knn}, \text{mem}\}} \gamma_f \, \eta_g^f \, {{}^{g}\!p^f_{ij}},
\end{align}}%
\revise{%
where ${{}^{g}\!p^f_{ij}}$ denotes either the standard nearest-neighbor probability or the membership-constrained probability. For each attribute $f$, the weights satisfy $\eta^f_{\mathrm{knn}}\ge0$, $\eta^f_{\mathrm{mem}}\ge0$, and $\eta^f_{\mathrm{knn}}+\eta^f_{\mathrm{mem}}=1$. Thus, $\eta^f_{\mathrm{knn}}$ controls how strongly the embedding follows attribute similarity independent of type, while $\eta^f_{\mathrm{mem}}$ controls how strongly it favors neighbors of the same type (see \autoref{fig:KNN_MEM}).
}
\revise{%
This formulation allows the latent space to reflect both local similarity and type-aware organization simultaneously. The mechanism is different from simply coloring points by type: type affects the graph used by the optimizer, but the attribute-specific visual neighborhoods remain explicit through $\gamma_f$.
}

\begin{figure}[h]%
\centering%
\scalebox{0.95}{%
\begin{tikzpicture}%
\begin{axis}[%
    width=0.49\textwidth,%
    height=0.25\textwidth,%
    xlabel={$x$},%
    ylabel={$g_c$},%
    legend style={at={(1.0,0.0)},anchor=south east},nodes={scale=0.5, transform shape},%
    domain=0:5,%
    samples=200,%
    thick,%
    xmin=0, xmax=5,%
    ymin=0, ymax=1,       
    clip=true,             
    enlargelimits=false,   
    axis lines=left,       
    grid=major,
    grid style={line width=0.2pt, draw=gray!40},  
    tick style={line width=0.2pt},
    font=\huge,
]
\addplot[oorange]   {x^2/(0.1+x^2)};
\addlegendentry{$c=0.1$}
\addplot[bblue]   {x^2/(0.5+x^2)};
\addlegendentry{$c=0.5$}
\addplot[ppurple]    {x^2/(1+x^2)};
\addlegendentry{$c=1~~\hphantom{.}$}
\addplot[ggreen] {x^2/(2+x^2)};
\addlegendentry{$c=2~~\hphantom{.}$}
\addplot[yyellow] {x^2/(5+x^2)};
\addlegendentry{$c=5~~\hphantom{.}$}
\addplot[lbblue] {x^2/(20+x^2)};
\addlegendentry{$c=20\hphantom{.}$}
\end{axis}%
\end{tikzpicture}%
}%
\caption{Function $g_c$ for different values of $c$.}
\label{Fig:g_c}
\end{figure}
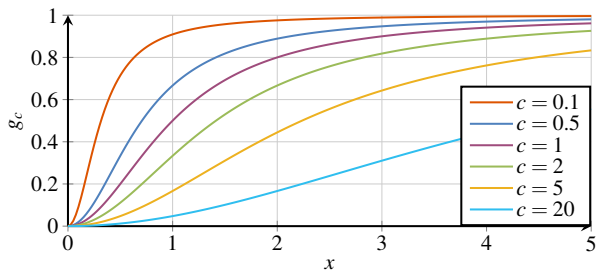

\paragraph*{Global Layout Energy.}
\revise{%
To obtain an understandable overview, we guide each point $\vy_i$ toward the vertex $\vh_i$ assigned to its watch type. The seven vertices are arranged as a regular heptagon only to provide a stable overview scaffold; their circular order is arbitrary and should not be interpreted as an ordinal relation between watch types. The layout term is therefore decayed during optimization so that it initializes and stabilizes the global structure without preventing local color and design neighborhoods from forming:
}
{\setlength{\abovedisplayskip}{4pt}
\setlength{\abovedisplayshortskip}{4pt}
\setlength{\belowdisplayskip}{4pt}
\setlength{\belowdisplayshortskip}{4pt}
\begin{align}
\label{Eq:LayoutLoss}
E_{\textrm{layout}} =
\frac{1}{2}\|\vy_i-\vh_i\|^2 - \frac{1}{2}c \log(c + \|\vy_i-\vh_i\|^2), \quad c>0,
\end{align}}%
with $\vh_i = \big( r \cos(\theta_i), \; r \sin(\theta_i) \big)$, with $\theta_i = \frac{2\pi i}{7}$, and where $r=5$ defines the radius of the heptagon.  
Analyzing the gradient gives: $\nabla E_{\textrm{layout}} = g_c(\|\vy_i-\vh_i\|^2)(\vy_i-\vh_i)$ with $g_c(x)=\frac{x^2}{c+x^2}$.
For small residuals \(x\), a Taylor expansion shows that $g_c(x)$ behaves like $g'_c(x)\approx\tfrac{x^2}{c}$, meaning that small errors are penalized more gently than with MSE, whose gradient grows linearly with the residual, see \autoref{Fig:g_c} for different values of $c$.
For large residuals, $g_c'(x)\approx 1-\tfrac{c}{x^2}$ grows essentially linearly, providing strong corrective forces when predictions are far from the target. 
This property ensures robustness to small fluctuations or noise. 

\subsection{Gradient-based Optimization}
\label{Sec:Optimization}
Combining the entropy in \autoref{Eq:CrossEntropy_Simple} with the layout loss in \autoref{Eq:LayoutLoss}, we obtain the final energy $E$:
{\setlength{\abovedisplayskip}{4pt}
\setlength{\abovedisplayshortskip}{4pt}
\setlength{\belowdisplayskip}{4pt}
\setlength{\belowdisplayshortskip}{4pt}
\begin{align}
    \label{Eq:E}
    E = E_{\textrm{entropy}} + \delta \cdot E_{\textrm{layout}}
\end{align}}%
where $\delta$ balances the two terms.
By default, we set $\delta=1$, and we refer to \autoref{Sec:ParameterEvaluation} for a parameter study. 
To maximize~\autoref{Eq:E}, we compute the energy gradient $\nabla E$,  cf.~\autoref{Eq:Attraction} and \autoref{Eq:Repulsion}:
{\setlength{\abovedisplayskip}{4pt}
\setlength{\abovedisplayshortskip}{4pt}
\setlength{\belowdisplayskip}{4pt}
\setlength{\belowdisplayshortskip}{4pt}
\begin{align}
    \nabla E &= \sum_{i\ne j}\Big[ a_{ij}P_{ij}(\vy_i-\vy_j)+r_{ij}(1-P_{ij})(\vy_i-\vy_j)\Big]- \nonumber\\
    &\qquad\qquad\delta\sum_i\frac{\lVert \vy_i - \vh_i \rVert^2}{c+\lVert \vy_i - \vh_i \rVert^2}(\vy_i-\vh_i).
\end{align}}%
Using gradient ascent on each low-dimensional point $\vy_i$, the energy $E$ is maximized with an adaptive learning rate $\alpha_n$: $ \vy_i^{(n+1)} \approx \vy_i^{(n)} + \alpha_n\cdot \nabla E$.

To prevent the layout term $E_{\textrm{layout}}$ from dominating the embedding throughout the optimization, we scale it with a factor $c$ that increases over time. Initially strong, the term gradually weakens during the UMAP optimization according to
$c=\frac{100}{\alpha_n^2}$, where $\alpha_n = 1-\frac{n}{N}$ and $n\in\{1,\ldots,N\}$ with $N=500$ iterations. 
Initially, attraction to the heptagon vertices is strong, establishing the global structure. As iterations increase, the layout term decays, allowing local refinement based on color and design neighborhoods captured by UMAP.

\begin{figure*}[t]%
    \centering%
    \includegraphics[width=0.32\linewidth]{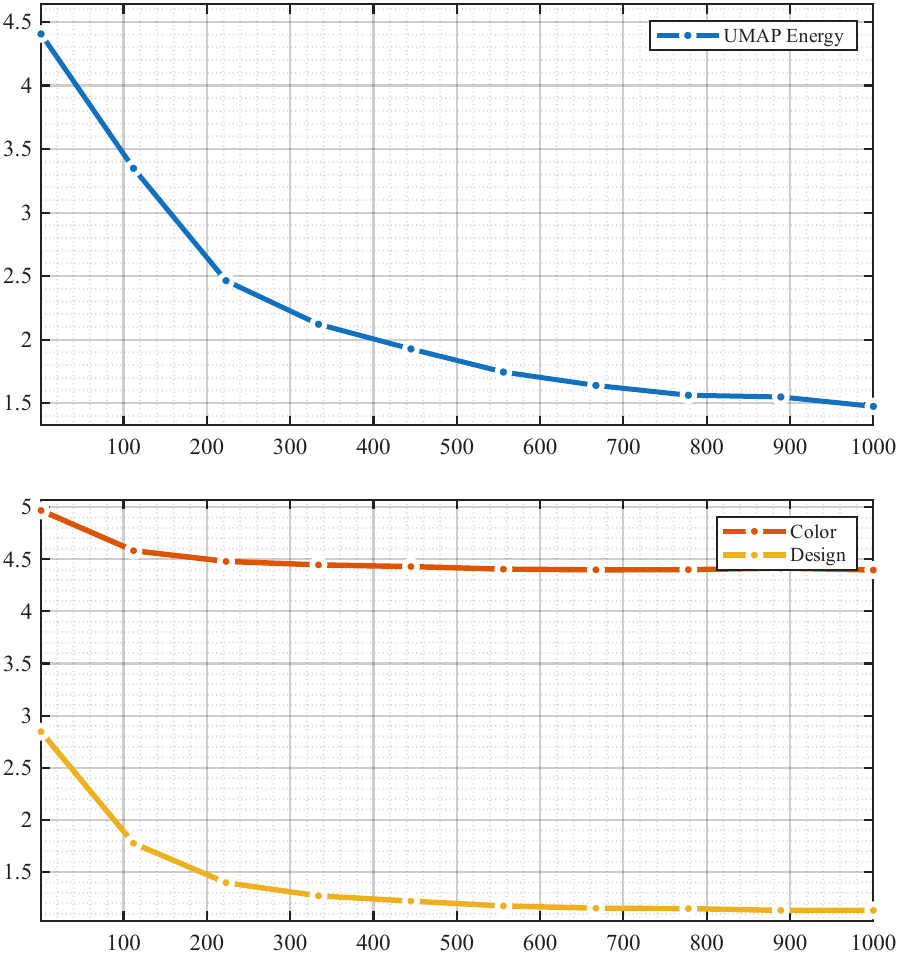}\hfill%
    \includegraphics[width=0.32\linewidth]{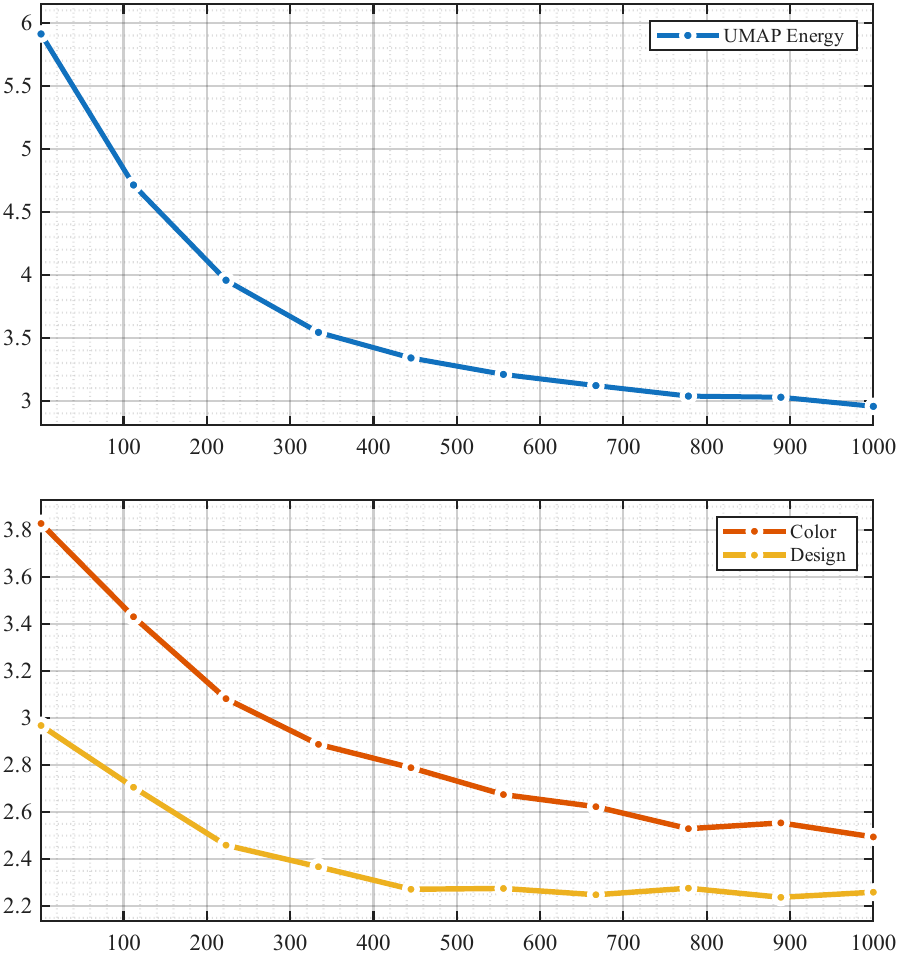}\hfill%
    \includegraphics[width=0.32\linewidth]{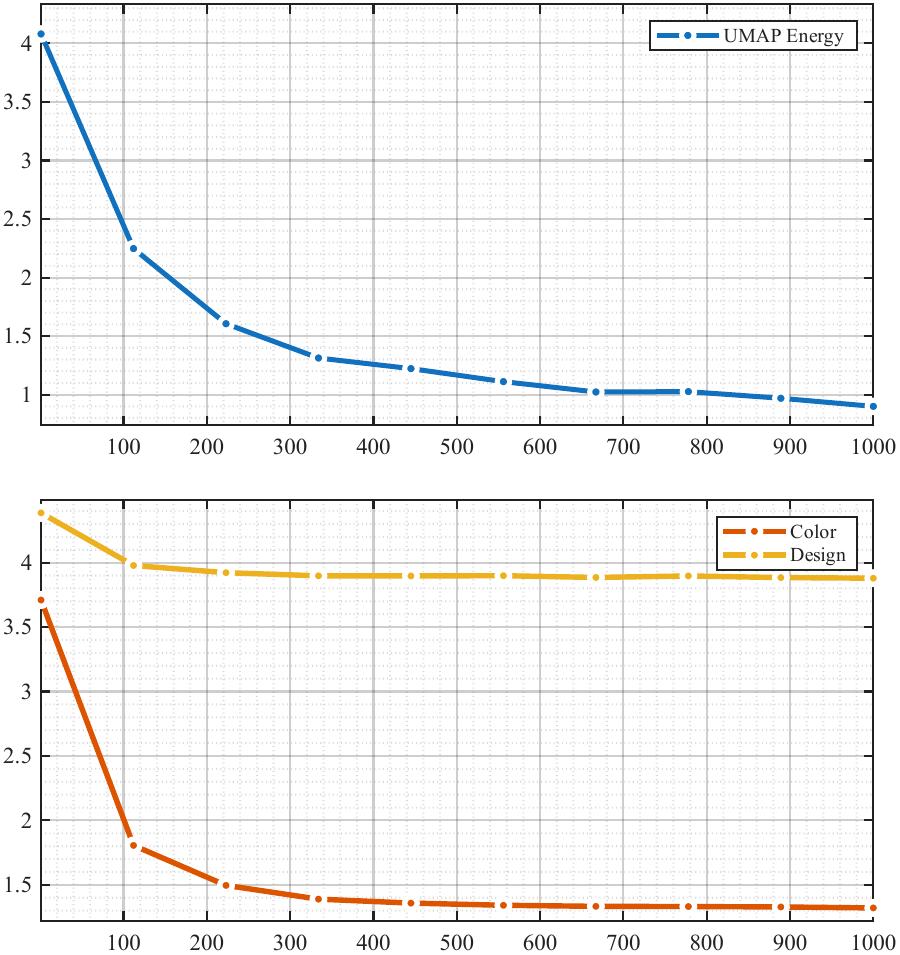}
    \begin{minipage}{0.32\linewidth}
        \centering \small $w_1=0,\,w_2=1$
    \end{minipage}\hfill%
    \begin{minipage}{0.32\linewidth}
        \centering \small $w_1=\nicefrac{1}{2},\,w_2=\nicefrac{1}{2}$
    \end{minipage}\hfill%
    \begin{minipage}{0.32\linewidth}
        \centering \small $w_1=1,\,w_2=0$
    \end{minipage}%
    
    \caption{\revise{Optimization behavior and layout-neighborhood quality for different attribute weights. The upper plots show the approximated UMAP loss over the optimization iterations. The lower plots show the mean color and design feature-space distances among nearest neighbors in the 2D layout; lower values indicate higher attribute similarity among layout neighbors. The loss decreases in all configurations, while the neighborhood-quality curves follow the chosen weighting: color is best preserved for ($w_1=1.0$), design is best preserved for ($w_1=0.0$), and the balanced setting preserves both attributes jointly.}}
    \label{Fig:energy_layout_w}
\end{figure*}

\revise{%
We evaluate the behavior of the embedding over the optimization process. 
We run the layout optimization with different attribute weights ($w_1\in{1.0,0.5,0.0}$), where ($w_1$) controls the influence of the color features and ($w_2=1-w_1$) controls the influence of the design features. 
For each setting, we record the approximated UMAP energy over 1000 iterations and additionally measure the attribute-space distances among the ($k=6$)-nearest neighbors in the resulting 2D layout. 
Lower attribute distances indicate that neighboring points in the layout are also more similar with respect to the corresponding high-dimensional attribute. 
Across all weight settings, the UMAP energy decreases rapidly during the first optimization phase and then gradually converges, indicating stable optimization behavior. 
The neighborhood-quality curves reflect the selected weights: for ($w_1=1.0$), the color-neighbor distance decreases strongly while the design distance remains comparatively high; for ($w_1=0.0$), the opposite behavior is observed, with a strong decrease in design distance and only limited improvement in color distance. 
For the balanced setting ($w_1=w_2=0.5$), both attribute distances decrease, showing that the layout preserves a compromise between color and design similarity. 
These results shown in \autoref{Fig:energy_layout_w} confirm that the optimization not only reduces the UMAP loss but also produces layouts whose local neighborhoods reflect the intended attribute weighting.
}


\section{Visualization System}
\label{sec:Visualization}
We designed an interactive visualization system to support the exploratory analysis of the latent watch space and to enable users to iteratively refine search strategies based on functional, aesthetic, and metadata attributes identified in our requirement analysis (\autoref{Sec:Requirement}). 

\paragraph*{Overview and Interaction Concept.}
The interface provides two coordinated views. The left view shows the 2D embedding from our multi-attribute UMAP, where each point represents a watch. Users navigate the space by zooming and panning to inspect both the global arrangement of watch types and the local similarities among watches.

When hovering over a point, the corresponding watch appears in a detail view on the right, showing the image and metadata such as brand, price, and case dimensions. This immediate feedback enables comparison of nearby watches while preserving context within the embedding.
Zooming and panning enable exploration of the latent space at different scales. Dense regions can be inspected to analyze subtle design variations, while zooming out reveals the global structure.
Together, these interactions support progressive exploration: users first identify relevant regions of the embedding and then examine nearby watches in detail to compare visually similar models.

\paragraph*{Heptagonal Layout for Watch Types (\autoref{W:type}).}

\begin{figure}[t]%
    \centering%
    \includegraphics[width=0.9\linewidth]{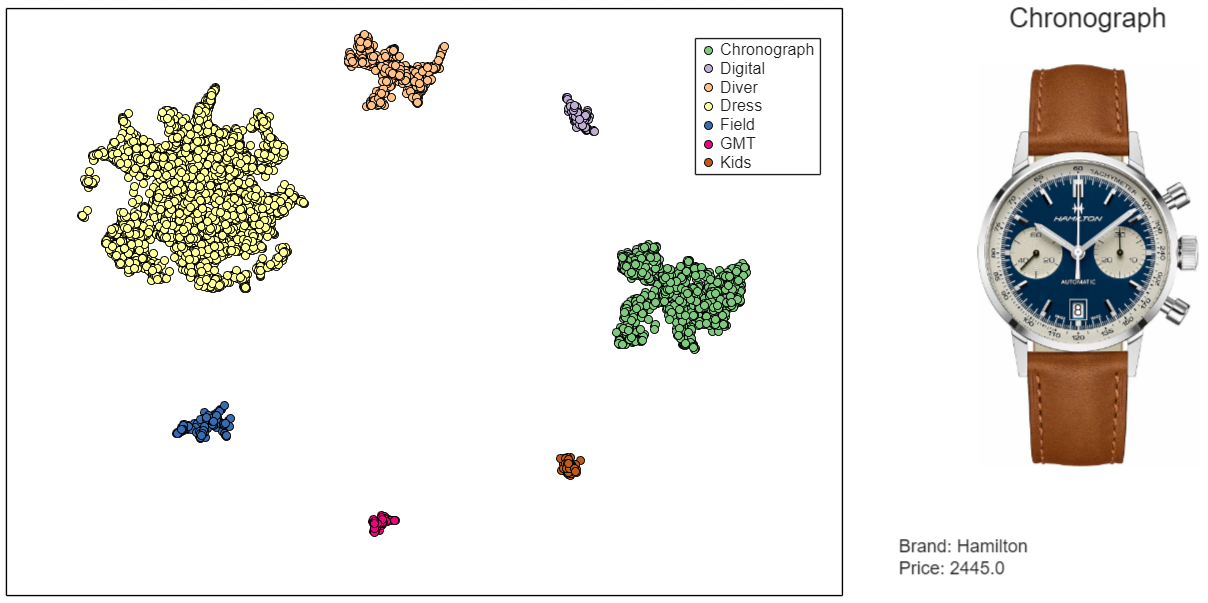}%
    \caption{The visualization supports watch exploration, with the multi-attribute UMAP on the left and details of the selected watch (e.g., brand, price) on the right.}%
    \label{fig:overview-va}%
\end{figure}

To support exploration by watch type (\autoref{W:type}), the latent space follows a heptagonal layout, see \autoref{fig:overview-va}. The seven vertices correspond to the watch types introduced in \autoref{sec:DomainAnalysis}. During the embedding optimization, a layout constraint attracts watches toward the vertex of their predicted type.
As a result, watches of the same type cluster around the corresponding heptagon vertex, allowing users to quickly identify functional groups and to focus on a specific watch type when navigating the embedding.

At the same time, the internal structure within each cluster reflects similarities in visual attributes such as color and dial design, supporting both categorical organization and exploration of visually related watches.
To avoid over-constraining the embedding, the strength of the layout term gradually decays during optimization, allowing the final arrangement to balance global structure and local similarity.

\paragraph*{Color-based Search (\autoref{W:color}).}
In the color-based search mode, users select a preferred dial color from an RGB palette. The selected color is converted to the perceptually uniform CIELAB color space to ensure that distances correspond more closely to perceived color similarity.

Using the 128 reference colors from the reference palette (\autoref{Sec:ColorFeatures}), we identify the eight palette colors closest to the user-selected color. With this, each watch is represented by a color attribute vector describing the contribution of these reference colors to its dial. For color-based queries, watches are retrieved if the contribution of any of the eight closest colors exceeds a user-defined threshold (10\,\% by default).
Matching watches are highlighted in the embedding, allowing users to easily identify watches with similar color characteristics. This mechanism enables users to discover watches with comparable color schemes even across different brands or watch types.

\paragraph*{Design-based Search (\autoref{W:design}).}
In addition to manual exploration, users can search the latent space for visually similar watches by uploading a reference image of a watch. After loading the image, the watch dial is segmented using the U-Net (\autoref{Sec:Segmentation}). If necessary, users may refine the segmentation mask interactively using brushing and erasing tools.
The segmented dial is then transformed into the same attribute representation used for the dataset. To insert the new watch into the latent space, we apply our adapted UMAP algorithm. We first determine the nearest neighbors of the query watch in the high-dimensional attribute space using the membership probability defined in~\autoref{Eq:MembershipProbability}. The new watch is then positioned close to these neighbors in the two-dimensional embedding. Unlike the original embedding process, where all points are jointly optimized, the structure of the existing latent space is preserved. The position of the inserted watch is refined using attractive forces only, cf.~\autoref{Eq:Attraction}, while omitting the repulsion term, cf.~\autoref{Eq:Repulsion}. This ensures that the new watch is placed near visually and semantically similar watches.

\begin{figure}[t]
    \centering
    \includegraphics[width=\linewidth]{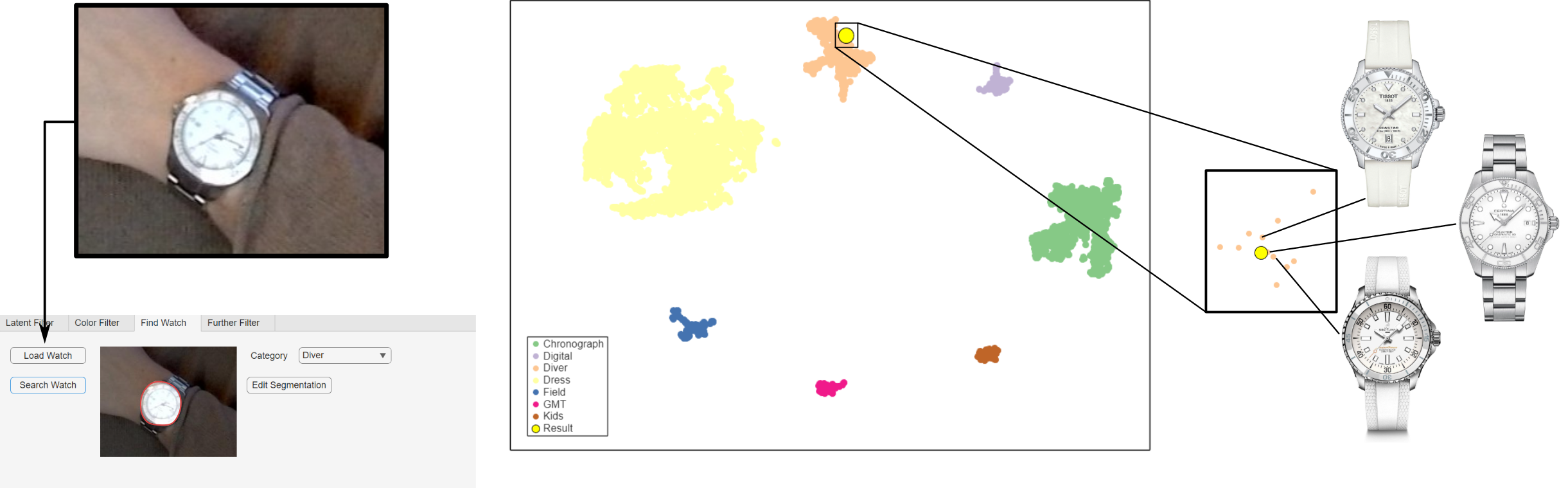}
    \caption{After loading a watch image, the dial is segmented and classified. The search highlights its position in the latent space and shows the three most similar watches, including the correct match (far right).    }
    \label{fig:SearchWatch}
\end{figure}

\begin{figure}[t]
    \centering
    \includegraphics[width=\linewidth]{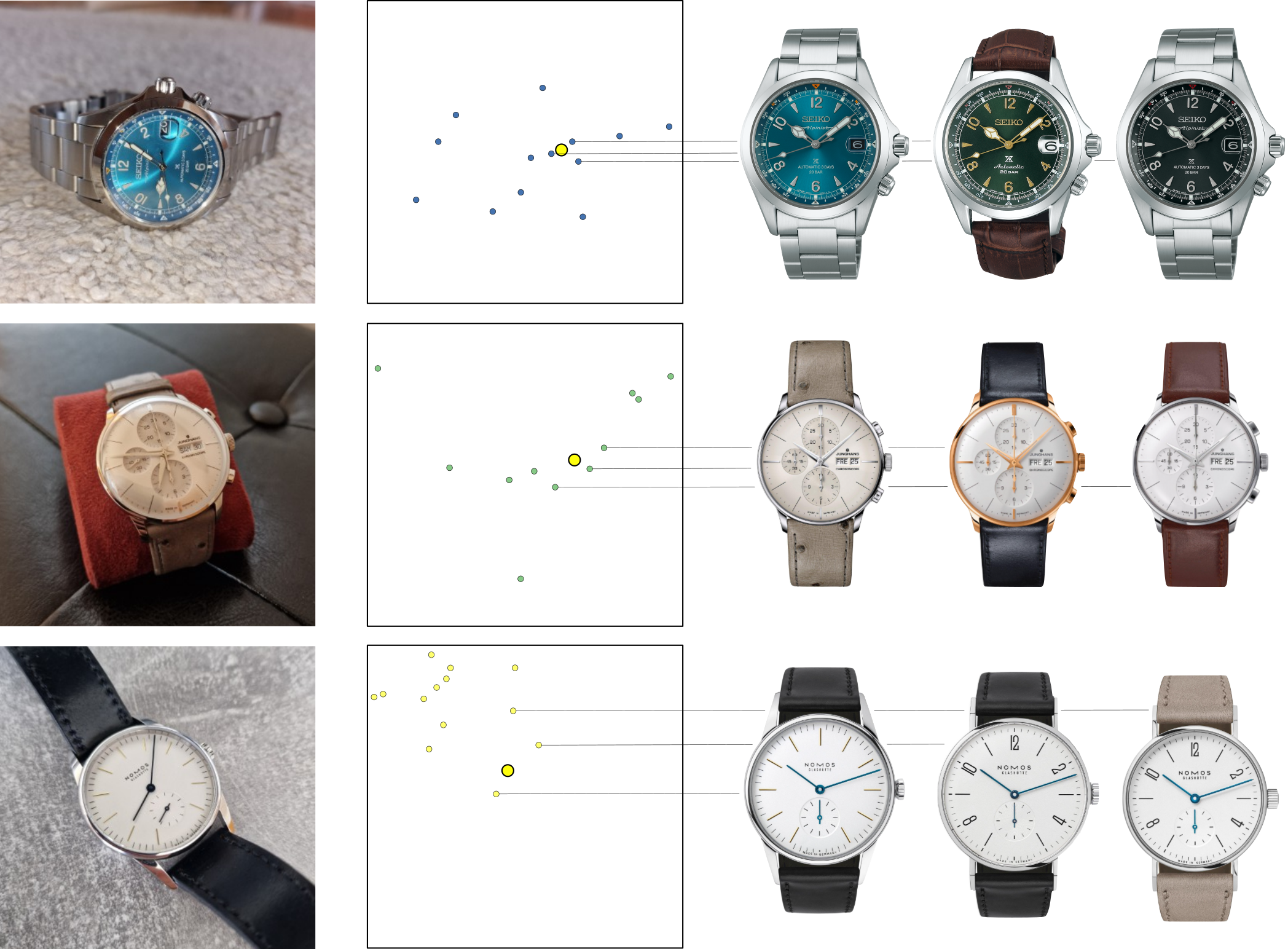}
    \caption{\revise{Example query watch image with corresponding nearest-neighbor results in the learned watch latent space. Retrieved watches are ordered from left to right by increasing latent-space distance.}   }
    \label{fig:SearchWatch_examples}
\end{figure}

The resulting position allows users to identify visually similar watches directly in the embedding. Nearby watches can then be inspected through the detail view to compare design elements such as dial layout, markers, or subdials. \autoref{fig:SearchWatch} \revise{and \autoref{fig:SearchWatch_examples} give examples for different input images.}%


\paragraph{\revise{Hybrid Exploration with Constraints (\autoref{W:hybrid}).}} 
\revise{%
In addition to visual attributes, the system provides metadata filters such as brand, price, and case dimensions. These filters allow users to restrict the displayed watches to items satisfying practical constraints. 
Combining metadata filters with latent-space visualization supports a typical exploration workflow: users may first identify visually interesting watches in the embedding and subsequently narrow the results using constraints such as price range, brand, or case size. This integration addresses \autoref{W:hybrid} by connecting open-ended visual discovery with conventional product filtering and detail inspection.}


\section{Evaluation}%
\label{Sec:Evaluation}~\hspace{-0.4em}%
\revise{%
We evaluate the approach in four complementary ways. First, we analyze parameters and runtime to show how the graph weights and layout term affect the embedding. Second, we quantitatively evaluate the search-by-example retrieval. Third, we report a qualitative pilot user study to understand how experts and novices use the interface. Fourth, we derive analytical insights and design implications. 
}

\begin{figure*}[t]%
    \centering%
    \includegraphics[width=0.24\linewidth]{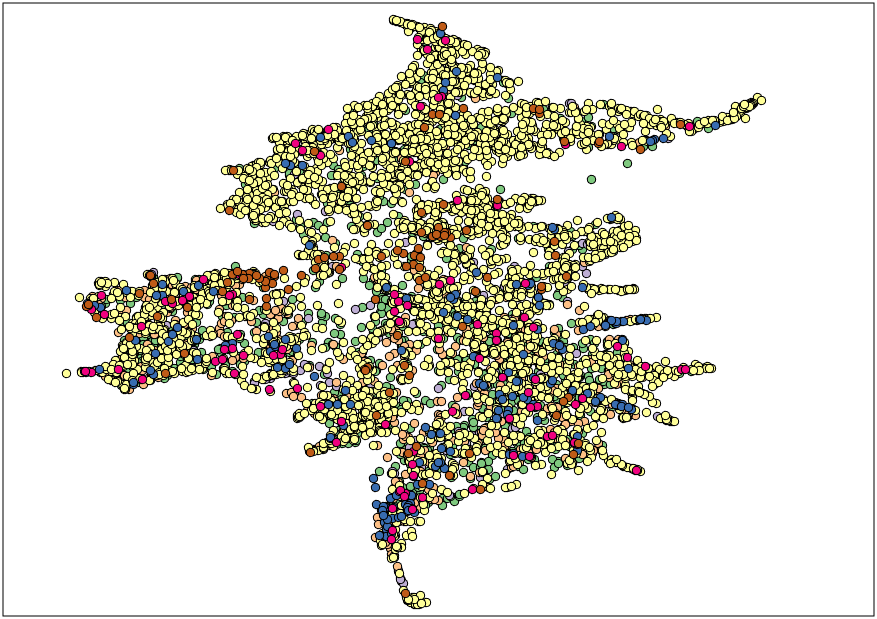}\hfill%
    \includegraphics[width=0.24\linewidth]{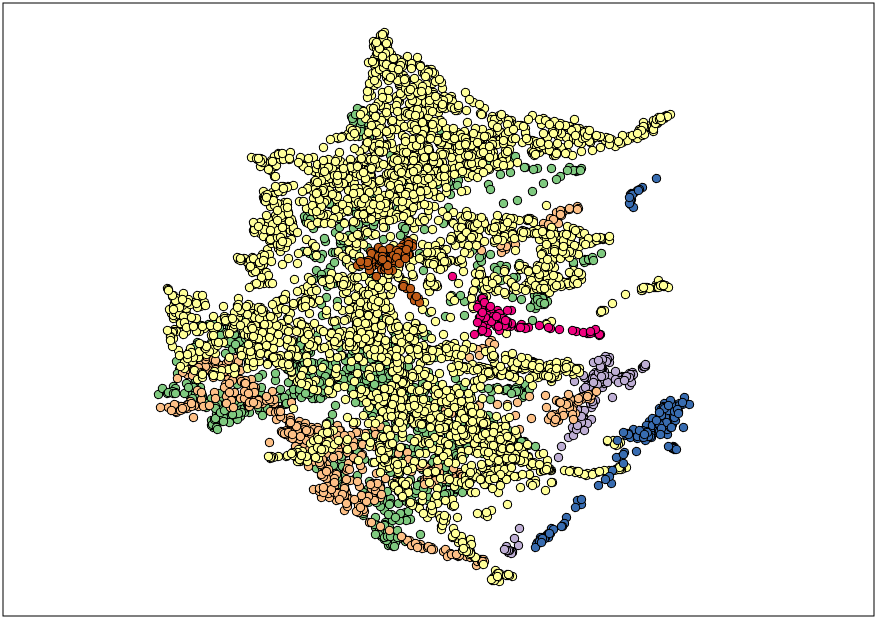}\hfill%
    \includegraphics[width=0.24\linewidth]{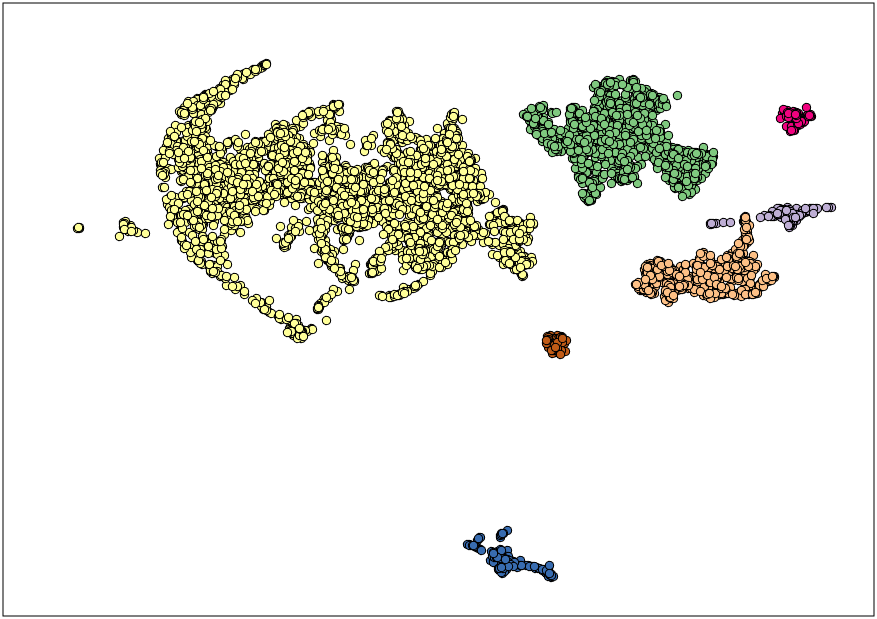}\hfill%
    \includegraphics[width=0.24\linewidth]{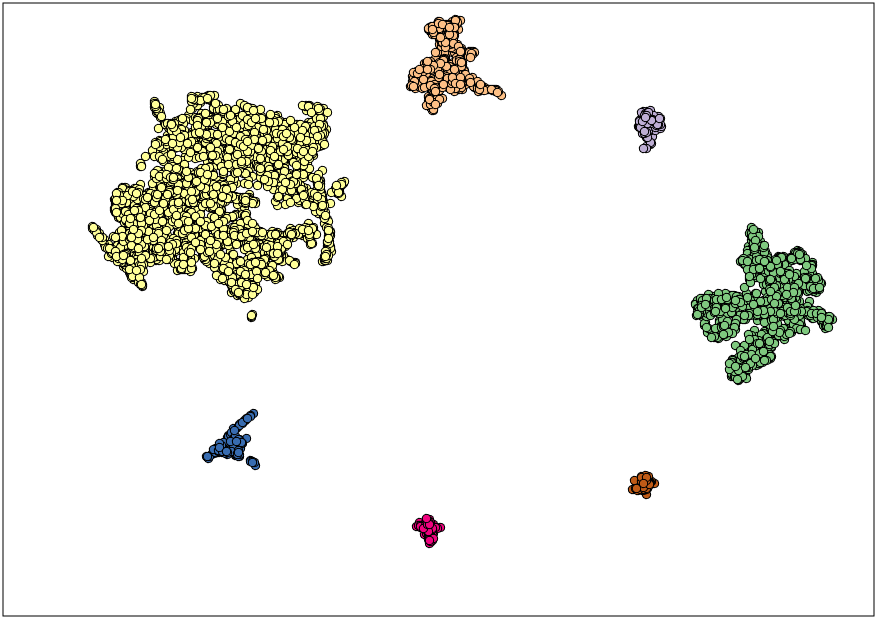}\\%
    \begin{minipage}{0.24\linewidth}
        \centering \small Color, $\eta_{knn}=1,~\overline N = 50\%$
    \end{minipage}\hfill%
    \begin{minipage}{0.24\linewidth}
        \centering \small Color, $\eta_{knn}=0.1,~\overline N = 75.39\%$
    \end{minipage}\hfill%
    \begin{minipage}{0.24\linewidth}
        \centering \small Color, $\eta_{knn}=0,~\overline N = 99.86\%$
    \end{minipage}%
    \begin{minipage}{0.24\linewidth}
        \centering \small Color, $\eta_{knn}=0,~\overline N = 100\%$
    \end{minipage}\\
    \includegraphics[width=0.24\linewidth]{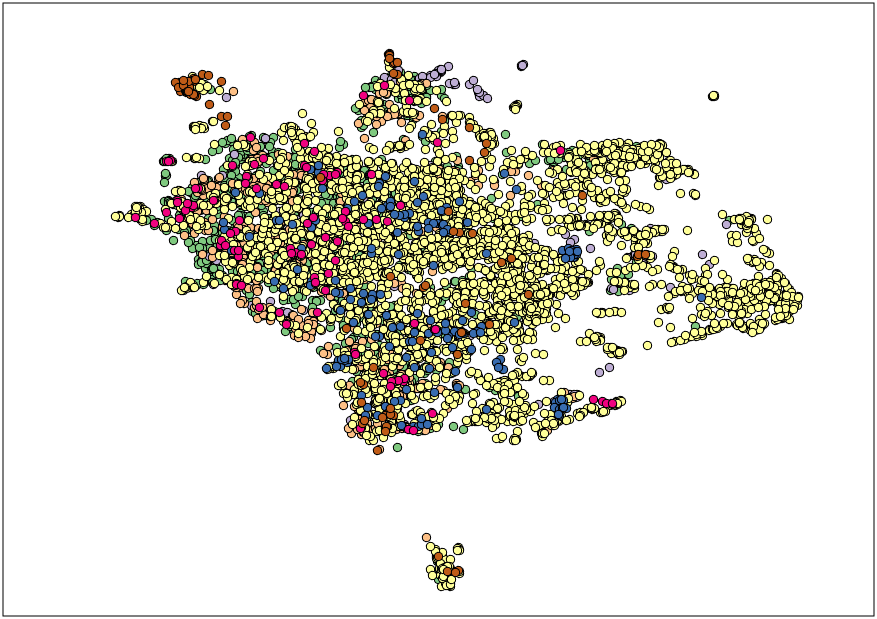}\hfill%
    \includegraphics[width=0.24\linewidth]{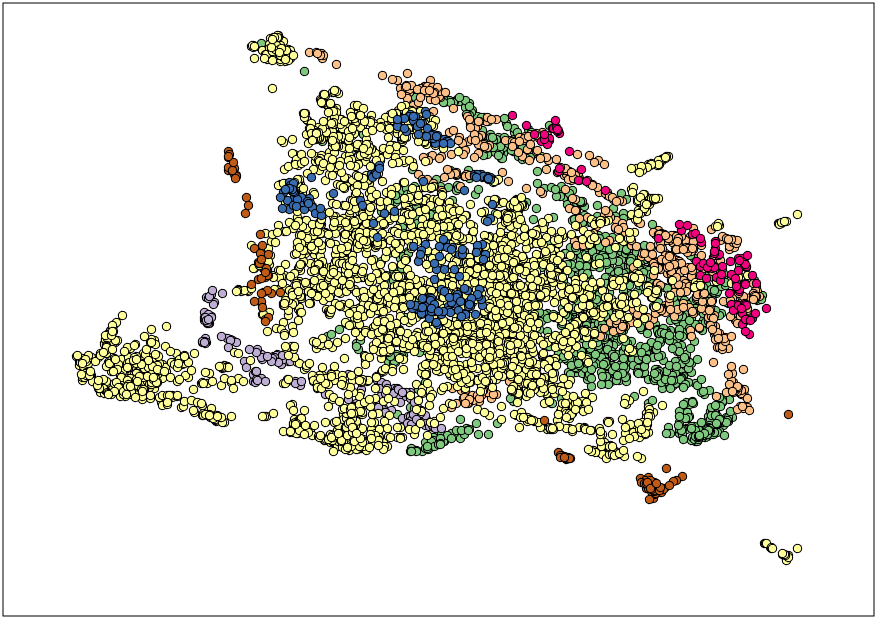}\hfill%
    \includegraphics[width=0.24\linewidth]{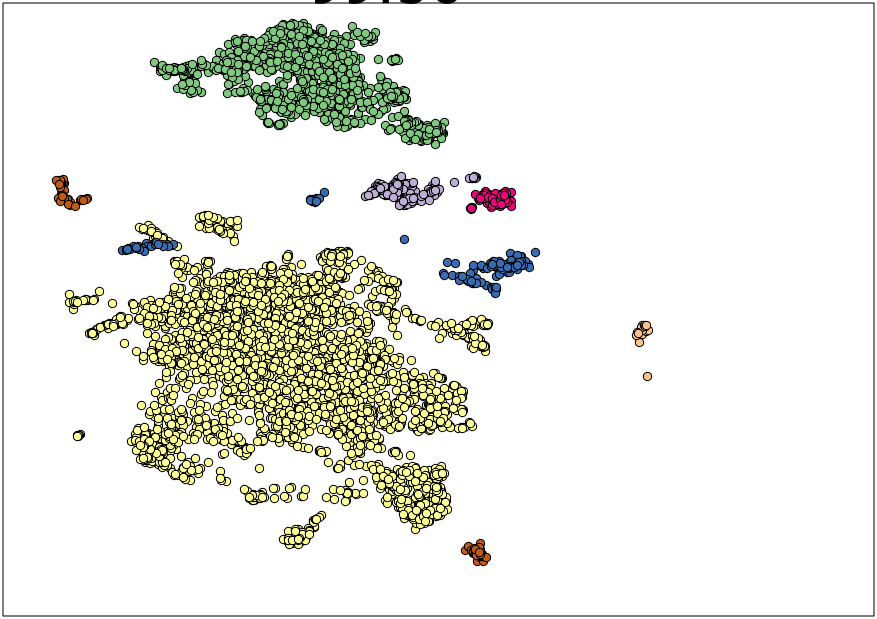}\hfill%
    \includegraphics[width=0.24\linewidth]{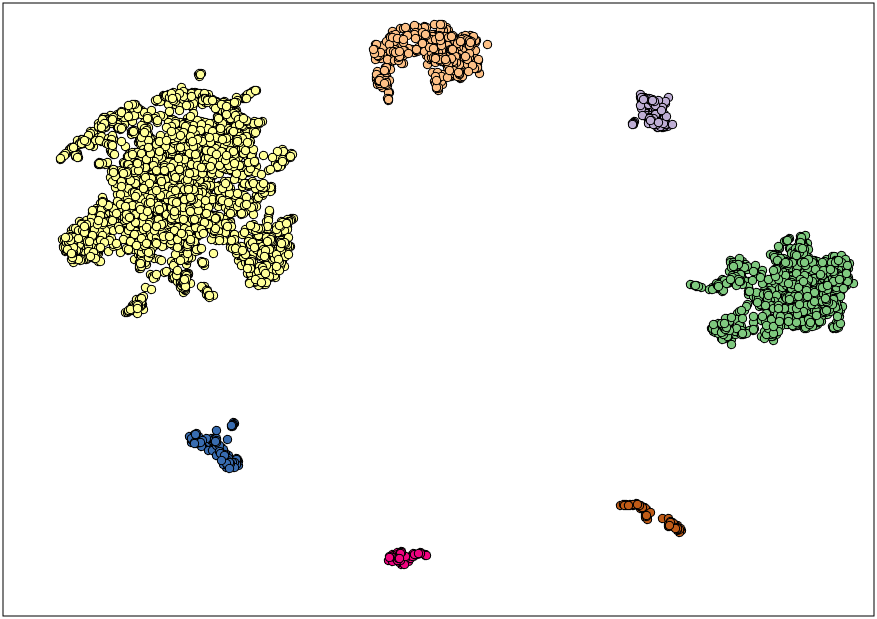}\\%
        \begin{minipage}{0.24\linewidth}
        \centering \small Design, $\eta_{knn}=1,~\overline N = 59.22\%$
    \end{minipage}\hfill%
    \begin{minipage}{0.24\linewidth}
        \centering \small Design, $\eta_{knn}=0.25,~\overline N = 73.47\%$
    \end{minipage}\hfill%
    \begin{minipage}{0.24\linewidth}
        \centering \small Design, $\eta_{knn}=0,~\overline N = 99.56\%$
    \end{minipage}%
    \begin{minipage}{0.24\linewidth}
        \centering \small Design, $\eta_{knn}=0,~\overline N = 100\%$
    \end{minipage}%
    \caption{Different parameter combinations for the latent space.  
For each configuration, we indicate whether the color attribute ($\gamma_{\text{col}}=1, \gamma_{\text{des}}=0$) or the design attribute ($\gamma_{\text{col}}=0, \gamma_{\text{des}}=1$) was used.  
We also report the contribution of the standard nearest neighbors ($\eta_{\text{knn}}$) and the membership-constrained neighbors ($\eta_{\text{mem}} = 1 - \eta_{\text{knn}}$), as well as the average number of points from the same class, $\overline{N}$.  
The layout term for the heptagon vertex arrangement was set to $\delta=0$ for all but the last column, where it was set to $\delta=1$.}
    \label{Fig:LatentParameter}
\end{figure*}

\subsection{Parameter and Performance Evaluation}
\label{Sec:ParameterEvaluation} 
To simplify parameter selection, we provide a set of \emph{presets} based on the requirements in \autoref{Sec:Requirement}, while experts may still adjust parameters for custom configurations. Depending on the analysis goal, the latent space can be organized by \emph{color} (\autoref{W:color}) or \emph{dial design} (\autoref{W:design}), and the heptagonal layout may be enforced ($\delta = 1$ in \autoref{Eq:E}) to reveal the seven \emph{watch types} (\autoref{W:type}). If the layout term is disabled ($\delta = 0$), only standard nearest neighbors are used ($\eta_{\text{knn}} = 1$, $\eta_{\text{mem}} = 0$), cf. \autoref{fig:LatentSpace_Color} and \autoref{fig:LatentSpace_HOG}.

\paragraph{Qualitative Parameter Study.}

\begin{figure*}[t]%
    \qquad\includegraphics[width=0.3\linewidth]{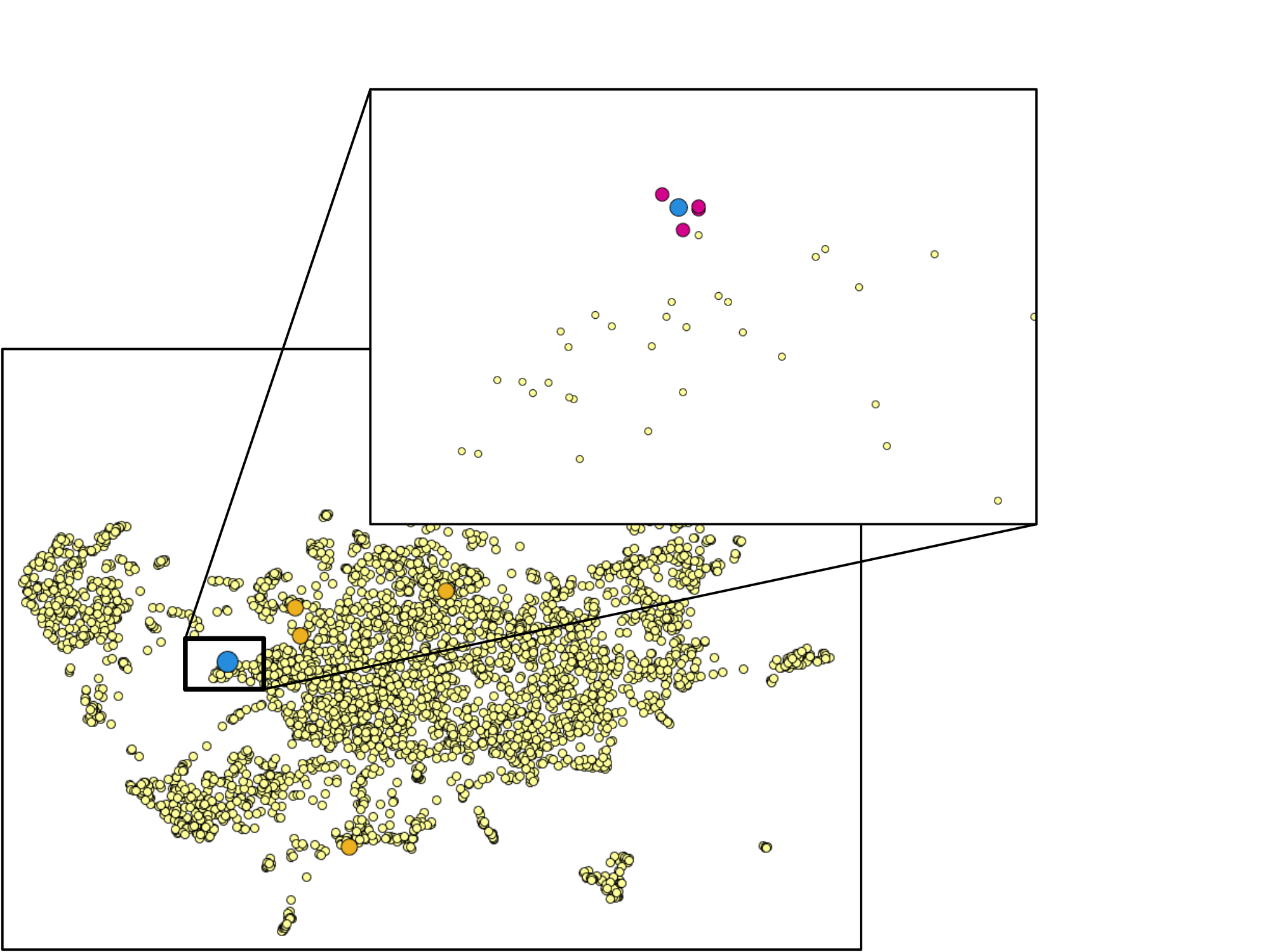}\quad%
    \includegraphics[width=0.3\linewidth]{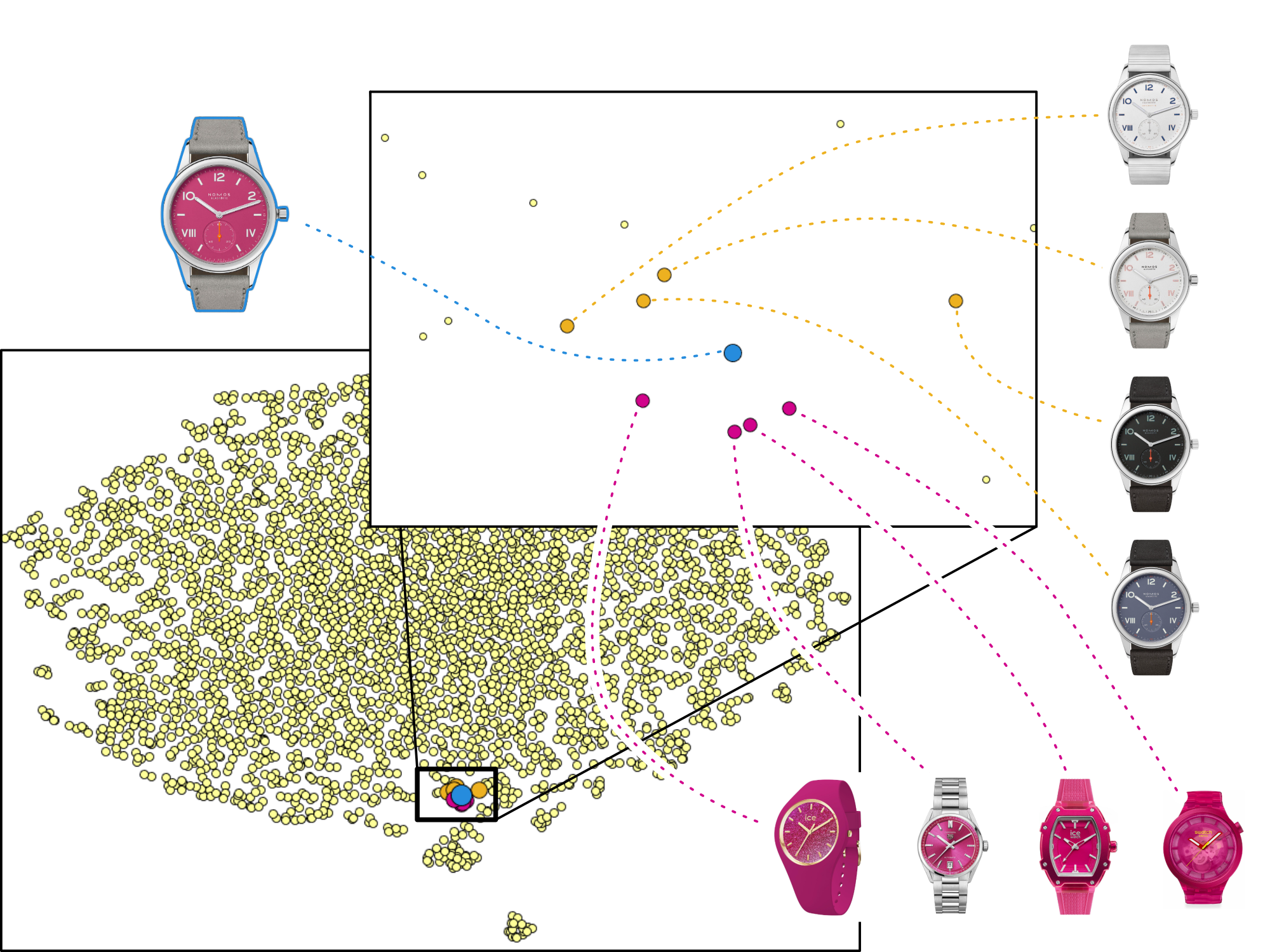}\qquad\qquad%
    \includegraphics[width=0.3\linewidth]{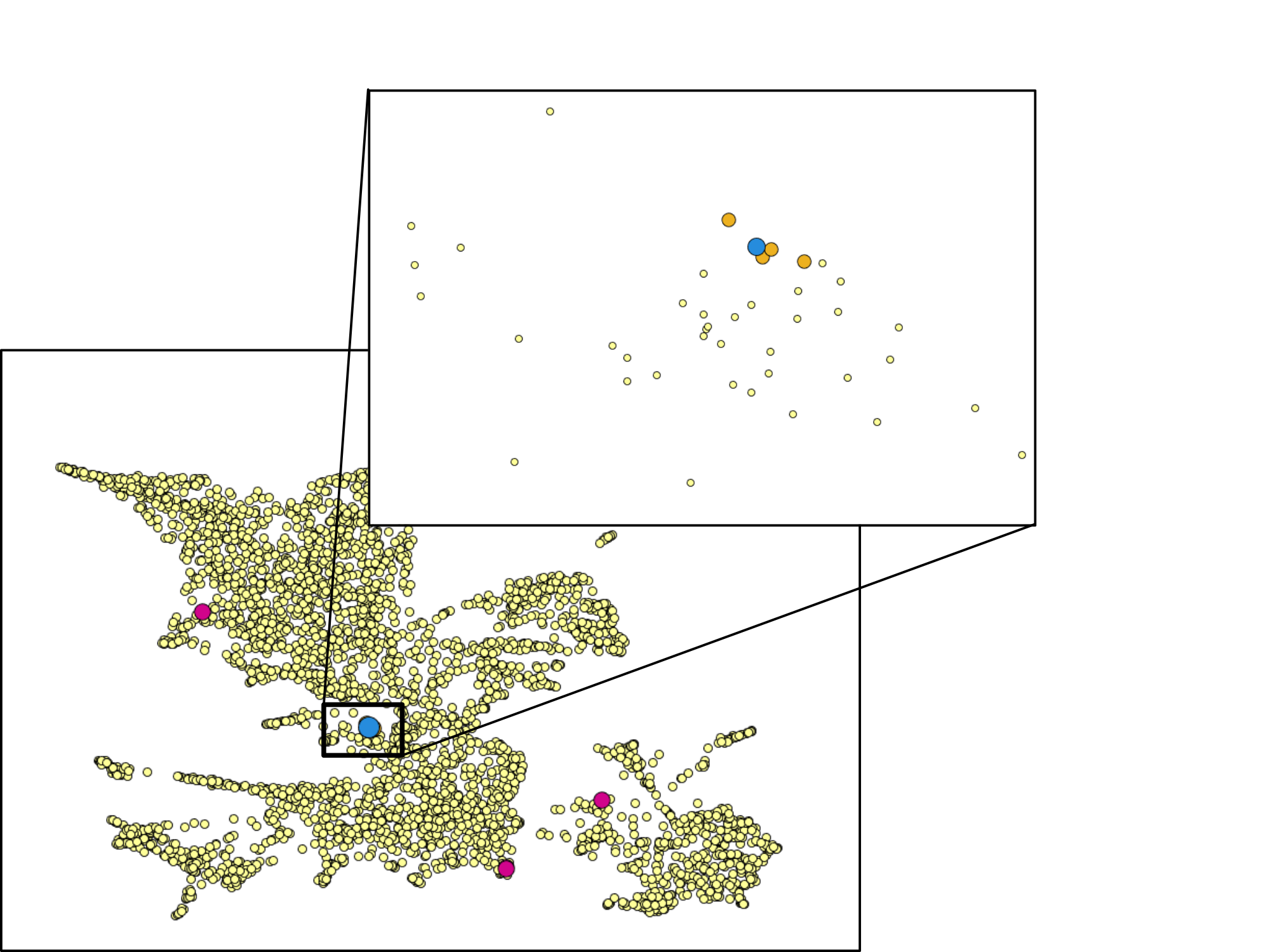}\\%
    \begin{minipage}{0.28\linewidth}
        \centering \small $\gamma_{\text{col}}=1, \gamma_{\text{des}}=0$
    \end{minipage}%
     \begin{minipage}{0.36\linewidth}
        \centering \small $\gamma_{\text{col}}= \gamma_{\text{des}}=0.5$
    \end{minipage}%
     \begin{minipage}{0.36\linewidth}
        \quad\centering $\gamma_{\text{col}}=0, \gamma_{\text{des}}=1$
    \end{minipage}%
    \caption{Mixed attribute space (center). The query point is shown in blue. As desired, nearby points are similar in both color and design.}%
    \label{fig:MA-UMAP-results}%
\end{figure*}

The attribute weights $\gamma_{\text{col}}$ and $\gamma_{\text{des}}$ control whether the embedding emphasizes color or design. For color-based organization, we use $\gamma_{\text{col}}=1$, $\gamma_{\text{des}}=0$, and for design-based organization, $\gamma_{\text{col}}=0$, $\gamma_{\text{des}}=1$. 
\autoref{Fig:LatentParameter} shows several parameter settings and their effect on the spatial arrangement and neighborhood composition. \revise{Here, we show the effect of membership neighborhoods and the heptagonal layout. We report a class-consistency-style neighborhood statistic $\overline{N}$, i.e., the average fraction of neighbors with the same watch type under different parameter settings.
As shown in \autoref{fig:MA-UMAP-results},} a balanced setting ($\gamma_{\text{col}}=\gamma_{\text{des}}=0.5$) yields a compromise between the two presets.

\paragraph{Quantitative Parameter Study.}

\revise{
We quantitatively measure how well the embedding reflects the influence of each attribute by calculating the Jaccard index $J_i$ between the high-dimensional neighborhood of attribute $i$ and the UMAP neighborhood:
\begin{align}
J_i = \frac{|N_{\text{attr}_i} \cap N_{\text{UMAP}}|}{|N_{\text{attr}_i} \cup N_{\text{UMAP}}|}
\end{align}
We then define the reconstruction error $E$ as the squared deviation between the observed neighborhood overlap and the assigned weights:
\begin{align}
E = \sum_{i=1}^{m} \delta_i\cdot(J_i - w_i)^2,~~~~\delta_i = \left\{
\begin{array}{ll}
0 & J_i \geq w_i \\
1 & \, \textrm{otherwise.} \\
\end{array}
\right.
\label{eq:error-metric}
\end{align}
The correction parameter $\delta_i$ ensures that errors are penalized only if there are not at least $w_i$ common neighbors in $J_i$. 
The results in \autoref{Tab:EvaluationNeighbors} show that the proposed graph-weighted embedding performs best for most mixed feature settings. In particular, it achieves the lowest neighborhood-preservation error across all settings in which both feature spaces contribute to the embedding objective. This supports the method's intended use case: the controlled combination of multiple feature-specific neighborhood graphs.

\begin{table}[h]
    \centering
    \caption{\revise{Quantitative assessment of the influence of each attribute, using error $E$ from Eq.~\eqref{eq:error-metric} for different weights and techniques. (Zero is best.)}}
    \begin{tabular}{cc|cc|c}
    \toprule
    \multicolumn{2}{c|}{Weights} & 
    \multicolumn{1}{c}{Joint (UMAP)} & 
    \multicolumn{1}{c}{N. joint (UMAP)} & 
    \multicolumn{1}{|c}{Our} \\\midrule
       $w_1=0$ & $w_2=1$ &   0.1605        & 0.1607            &  0.1595 \\
       $w_1=\nicefrac{1}{4}$ & $w_2=\nicefrac{3}{4}$ &   0.4501        & 0.3510            &  \textbf{0.0635} \\
       $w_1=\nicefrac{1}{2}$ &  $w_2=\nicefrac{1}{2}$ &   0.3482        & 0.3656            &  \textbf{0.1220} \\
       $w_1=\nicefrac{3}{4}$ & $w_2=\nicefrac{1}{4}$ &   0.4502        & 0.4395            &  \textbf{0.1744} \\
       $w_1=1$ & $w_2=0$ &   0.3464        & 0.3393            &  0.3362 \\
       \bottomrule
    \end{tabular}
    \label{Tab:EvaluationNeighbors}
\end{table}

Further justification for the metric in \autoref{eq:error-metric} and a study similar to \autoref{Tab:EvaluationNeighbors} for the synthetic motivational example in \autoref{fig:motivationalExample} can be found in the supplemental material.
}

\paragraph{Performance.}
Performance was measured on an Intel(R) Core(TM) i7-14700KF CPU (3.40 GHz), 48 GB RAM, and an NVIDIA GeForce RTX 4080 SUPER GPU.
Dial segmentation (\autoref{Sec:Segmentation}) required $\approx$ $0.043\,s$, watch type classification $0.319\,s$ (\autoref{Sec:Category}), and the UMAP optimization (500 iterations) $0.543\,s$ with an additional $0.074\,s$ to compute the fuzzy simplicial sets. Overall, our extensions introduce negligible computational overhead compared to standard UMAP.

\revise{\subsection{Search-by-Example Retrieval Evaluation}
The search-by-example interaction in \autoref{sec:Visualization} allows users to insert a
user-provided watch image into the fixed latent space and inspect nearby
watches as visually similar alternatives. 
To complement the qualitative example in \autoref{fig:SearchWatch}, we added a controlled retrieval evaluation that quantifies how well the out-of-sample insertion places a query close to its corresponding original watch.
We treat the existing embedding as a gallery and use gallery images as query images. 
For each query, the corresponding original gallery image is considered the correct match and remains part of the gallery. 
The query image is processed with the same descriptor pipeline as in the interactive system: the dial mask is estimated with the U-Net segmentation model and the visual descriptors are extracted. 
We evaluate three attribute settings. 
The \emph{color-only} setting uses the dial color histogram, the \emph{design-only} setting uses the HOG descriptor, and the \emph{mixed} setting combines color and HOG with equal weights. 
For each setting, a fixed gallery embedding is computed once and remains unchanged during the evaluation. 
Each query is then inserted into the corresponding fixed embedding by identifying its nearest neighbors in the selected high-dimensional attribute space and placing the query in the embedding space. 
To evaluate robustness, each query image is modified by a randomly sampled perturbation before descriptor extraction. 
The perturbation combines geometric and photometric changes intended to approximate variations in user-provided images. 
Specifically, we apply an in-plane rotation sampled uniformly from \([-3^\circ, 3^\circ]\), isotropic scaling from \([0.97, 1.03]\), and horizontal and vertical translations of up to \(1.5\%\) of the image width or height. 
In addition, brightness is scaled by a factor sampled from \([0.90, 1.10]\), contrast by a factor sampled from \([0.95, 1.05]\), Gaussian blur with \(\sigma \in [0, 0.30]\) pixels is applied, and additive Gaussian image noise with standard deviation \(\sigma=0.002\) is added in normalized intensity space. No occlusion is applied.
After insertion, gallery watches are ranked by Euclidean distance to the inserted query point in the 2D latent space. 
We report Top-1 accuracy (R@1), Recall@5, Recall@10, and the mean reciprocal rank (MRR) of the corresponding original watch. 

The results in \autoref{tab:search_by_example_retrieval} show that all three attribute settings reliably place perturbed query images close to their corresponding original watches.
The design-only setting performs best overall. 
This indicates that the HOG-based design descriptor provides a highly stable basis for the search-by-example interaction under mild geometric and photometric perturbations. 
The color-only setting also performs well, 
showing that the dial color descriptor remains
discriminative for perturbed queries. 
While combining color and design provides a balanced representation for exploration, the fixed equal weighting does not outperform the design-only setting in this controlled retrieval task.
}

\begin{table}[h]
    \centering
    \caption{\revise{Controlled search-by-example retrieval evaluation with
    out-of-sample insertion under query perturbations.
    For each attribute setting, all perturbed query instances are
    inserted into a fixed gallery embedding. }}
    \label{tab:search_by_example_retrieval}
    \begin{tabular}{l|rrr|r}
        \toprule
        Attribute setting
        & R@1
        & R@5
        & R@10
        & MRR         \\
        \midrule
        Color-only
        & 0.782 & 0.837 & 0.844 & 0.809  \\
        Design-only
        & 0.882 & 0.960 & 0.970 & 0.915  \\
        Mixed
        & 0.830 & 0.850 & 0.858 & 0.840 \\
        \bottomrule
    \end{tabular}
\end{table}

\subsection{User Evaluation}
\revise{%
The user evaluation focuses on whether the system supports the intended exploration workflow: global orientation, local comparison, filtering, and discovery. We recruited both experts and novices because the system targets two complementary forms of use. Experts can judge whether neighborhoods and type regions match domain knowledge, while novices reveal whether the map is understandable without extensive horological expertise. We do not use the two groups for inferential statistics; the sample size is too small for that purpose. Instead, we analyze task completion, descriptive ratings, and open-ended comments to identify recurring usage patterns and failure cases.
}

\subsubsection{Participants and Procedure}
We recruited ten participants: four watch experts and six novices. The expert group included one female and three males, and the novice group two females and four males, aged 23–67 years (\emph{M} = 40.0, \emph{SD} = 13.5). Before the study, participants completed a short questionnaire on their familiarity with watches and online product exploration.
The study was conducted on a desktop system where participants could zoom, pan, hover for details, and apply filters by, e.g., price, case size, or brand. Each session lasted about 60 minutes and included an introduction, five exploration tasks, and a post-study questionnaire (see suppl. material).
\revise{%
All participants were informed about the purpose and procedure of the study and provided informed consent before taking part.}

\subsubsection{Tasks}
Our participants completed the following five tasks:
\begin{itemize} [topsep=2pt]
\item[\textbf{T1}]\textbf{Global Organization:} Identify three watches of a specific type (e.g., divers) located close in the latent space.
\textit{Goal:} Assess whether global structure and semantic grouping are recognizable.

\item[\textbf{T2}] \textbf{Local Refinement:} Find watches that share the same type but differ in dial color or design.  
\textit{Goal:} Evaluate local clustering by aesthetic similarity.

\item[\textbf{T3}] \textbf{Type Comparison:} Compare two types (e.g., divers and dress watches) and describe visual differences.  
\textit{Goal:} Assess the interpretability of inter-cluster relationships.

\item[\textbf{T4}] \textbf{Attribute Filtering:} Apply metadata filters (e.g., price $< \$2000$, case size $<40$ mm) to find relevant watches.
\textit{Goal:} \revise{Evaluate how non-visual attributes support targeted exploration and whether users can combine visual neighborhoods with practical constraints (\autoref{W:hybrid}).}

\item[\textbf{T5}] \textbf{Discovery:} Identify one new or unexpected watch that the participant finds visually appealing. 
\textit{Goal:} Measure the system’s ability to support the discovery of consumer-relevant watches.

\end{itemize}

\subsubsection{Study Criteria and Questionnaire}

The questionnaire assessed the functional performance and the user experience of our tool, regarding the following four criteria:

\begin{enumerate} [topsep=2pt]
\item \textbf{Effectiveness (5-point Likert):} Assess how efficiently and accurately users can identify and compare watches within the latent space.  
Example: “The visualization helped me find visually or semantically similar watches easily.”
\item \textbf{Interpretability (5-point Likert):} Evaluate how well users understand the spatial organization and attribute-driven structure of the visualization.  
Example: “I understood why certain watches were positioned close to each other.”
\item \textbf{Usability (SUS, 5-point Likert, 0--100 score):} Measure perceived ease of use, learnability, and interaction quality using the standardized \textit{System Usability Scale (SUS)}~\cite{brooke1996sus}.  
Example: “I thought the visualization was easy to use.”
\item \textbf{User Experience (UEQ-S, 7-point bipolar, –3 to +3):} Capture overall satisfaction, engagement, and emotional response using the short \textit{User Experience Questionnaire (UEQ-S)}~\cite{hinderks2017}, covering \textit{Pragmatic} and \textit{Hedonic Quality}.  
Example: “Using the visualization felt exciting/boring.”
\end{enumerate}

In addition, participants shared open-ended feedback on helpful features, sources of confusion, and suggestions for improvements.

\subsubsection{Study Evaluation}

All participants completed the tasks, providing initial evidence that the system supports the intended exploration workflow. Differences between experts and novices emerged in efficiency, usability, and exploration strategies.

\paragraph{Task Performance.}  
Experts completed most tasks with high accuracy and efficiency, especially those involving type identification and visual comparison (T1–T3). 
They quickly grasped the latent-space structure and used spatial proximity to infer design or color similarities. 
On average, experts achieved 94\,\% accuracy with a mean completion time of 3.2\,minutes per task (\emph{SD} = 0.8). 
Minor errors occurred when visually similar models from different brands overlapped in dense regions, leading to occasional misclassification of design variants.
Novices also performed well, achieving an average accuracy of 84\,\% with a mean completion time of 4.3\,minutes per task (\emph{SD} = 1.2). 
Performance declined slightly on tasks requiring the interpretation of spatial relationships (T2, T4), in which proximity was sometimes mistaken for exact functional similarity. 
Some novices initially felt uncertain when distinguishing clusters or navigating while zooming and filtering, but their confidence and accuracy improved noticeably over time.

\paragraph{Effectiveness and Interpretability.}  
Experts rated the \textit{effectiveness} highly (\emph{M} = 4.6, \emph{SD} = 0.5), praising the clear spatial grouping of watch types and the ease of comparing similar models. 
Novices rated the visualization's effectiveness slightly lower (\emph{M} = 4.0, \emph{SD} = 0.7) but valued how it revealed aesthetically related designs and the variety of watch types. 
Lower scores were mainly due to uncertainty about how filters affect the layout and difficulty in noticing subtle design nuances within clusters.
Experts rated \textit{Interpretability} positively (\emph{M} = 4.4, \emph{SD} = 0.5), characterizing the spatial organization as intuitive and consistent with domain knowledge. 
They noted that distances generally reflected perceived similarity and could often infer which attributes influenced clustering. 
Novices rated interpretability slightly lower (\emph{M} = 3.9, \emph{SD} = 0.8), reporting occasional confusion about proximity relations.  
Some users initially struggled to understand the layout but improved once they recognized how color and design shaped local groupings. 
Both groups suggested adding further interactive explanations or legends.

\paragraph{Usability (SUS).}  
Experts rated the system’s usability highly (SUS \emph{M} = 85.3), praising its clear interface, smooth interactions, and logical filter arrangement.  
Novices gave slightly lower ratings (\emph{M} = 73.8) and reported minor challenges during exploration.
The most common issue was the \emph{zoom-to-detail transition}, where rapid zooming caused brief disorientation or loss of context. 
Some novices reported brief delays when applying multiple filters, while a few experts suggested clearer feedback to indicate which filters were active. 
Regardless, participants agreed that the visualization was easy to learn, offered sufficient control, and provided a coherent, responsive experience. 
Taken together, the SUS scores suggest that the interface was generally usable and learnable, including for first-time users.

\paragraph{User Experience (UEQ-S).}  
\revise{%
UEQ-S results were strongly positive for both groups, indicating that the visualization was perceived as effective and enjoyable. Experts rated \textit{Pragmatic Quality} highest (\emph{M} = 2.4, \emph{SD} = 0.5), highlighting clarity, precision, and control during exploration, as well as the ability to compare watch types and designs without losing context. Novices rated this dimension lower (\emph{M} = 1.4, \emph{SD} = 1.1), reporting occasional difficulty interpreting distances and understanding the interaction between color and type, especially when multiple filters were active.
For \textit{Hedonic Quality}, both groups reported similarly positive ratings (experts: \emph{M} = 2.7, \emph{SD} = 0.4; novices: \emph{M} = 2.0, \emph{SD} = 0.9). Participants described the interface as aesthetically pleasing, engaging, and enjoyable to explore. Experts appreciated insights into stylistic relationships, while novices valued discovering designs and types they would not have explored otherwise. Some suggested adding guidance, such as tooltips, to improve navigation.
}

\paragraph{\revise{Hybrid Exploration with Constraints (\autoref{W:hybrid}).}}
\revise{%
Task T4 provides initial evidence for \autoref{W:hybrid}. Participants were able to combine the latent-space overview with metadata constraints such as price and case size, indicating that the system supports a hybrid exploration workflow. Several users first identified visually interesting regions in the embedding and then refined their search using practical constraints, which corresponds to the intended transition from visual discovery to targeted product comparison. However, this result should be interpreted as qualitative evidence rather than a complete validation of constraint-based decision-making.}

\subsection{Analytical Insights Enabled by the Visualization}

Beyond performance and usability, the study revealed analytical insights enabled by the visualization, showing how the latent-space embedding, heptagonal layout, and interaction support analysis and discovery in this domain.

\paragraph{Discovery of Previously Unknown Designs.} Participants frequently reported discovering previously unknown watch models, enabled by the spatial organization and navigation interactions such as zooming and panning. After identifying a watch of interest, they explored nearby regions to inspect visually related designs.

The hover-based detail view further supported this process by allowing quick inspection of nearby watches without interrupting spatial exploration. Several participants noted that this workflow enabled intuitive browsing of stylistically related watches. One novice participant commented that the system allowed them to “find watches that look very similar to ones I like, but from brands I did not know before.” An expert similarly remarked that “once I find a watch I like, I can just look around it and quickly see related designs.”

\paragraph{Revealing Stylistic Relationships within Watch Types.} The global heptagonal layout helped participants understand the functional structure of the watch collection, as each vertex corresponds to a watch type. This allowed users to quickly identify regions associated with chronographs, diver watches, or dress watches. Within these regions, the latent-space embedding organizes watches according to visual attributes such as dial design and color.

The combination of global categorical structure and local visual similarity enabled participants to identify stylistic relationships within watch types. Clusters often reflected similar dial layouts or design elements, allowing experts to compare watches with similar configurations of subdials, markers, or textures. One expert noted that “you can clearly see which chronographs follow a similar dial layout,” illustrating how the visualization revealed stylistic patterns that would otherwise require manual comparison across many products.

\paragraph{Identifying Visual Similarity Across Brands.} The embedding enabled participants to identify visually similar watches across different brands. Because the latent space is organized primarily by visual attributes rather than brand metadata, watches with similar color palettes or dial structures often appear close together, even when produced by different manufacturers.
Participants used spatial navigation and the hover-based detail view to inspect such relationships. In several cases, users noticed watches from different brands positioned near each other due to shared visual characteristics. One participant remarked that “normally I would search within a single brand, but here I can see watches from different brands that have almost the same style.” This capability was described as particularly useful for discovering alternatives.

\paragraph{Supporting Visual Exploration Strategies.} The visualization encouraged exploration strategies combining global orientation with local inspection. Participants typically began by examining the overall embedding, where the heptagonal layout provides an overview of watch types, and then navigated to regions of interest using zooming and panning. Within these regions, they used the hover interaction and detail panel to inspect individual watches and compare nearby designs. Some users further refined their search using metadata filters or by inserting a reference image to locate visually similar watches.

Participants explicitly described this workflow. One noted that “I first look at the big clusters and then zoom into interesting areas,” while another summarized it as “starting with the overall map and then exploring nearby designs.” These observations suggest that the combination of global layout structure, local similarity, and interactive inspection supports progressive exploration of the design space.


\section{Design Implications for Latent Spaces}
The design and evaluation of our visualization system reveal implications for visual analysis tools operating on heterogeneous datasets. Many real-world collections combine categorical attributes, visual descriptors, and metadata that jointly define similarity relationships. Effective visualization of such data requires balancing semantic structure, visual similarity, and user-driven exploration. Based on our system design and user study findings, we derive implications that may inform visualization systems for domains such as fashion collections, product catalogs, industrial design archives, or image repositories.

\paragraph{Balancing Global Structure and Local Similarity.} 
A key challenge in visualizing heterogeneous collections is balancing global semantic organization with local visual similarity. Categorical attributes define high-level groupings, while visual descriptors capture finer differences. Treating all attributes equally during dimensionality reduction can cause these relationships to interfere and produce hard-to-interpret embeddings.
Our approach separates global structure from local similarity. The global layout is guided by categorical information (watch type), while local neighborhoods are determined by visual attributes such as color and dial design. This allows users to quickly identify semantic regions and explore detailed variations. More generally, visualizations of heterogeneous collections benefit from distinguishing attributes that define global structure from those shaping local similarity.

\paragraph{Supporting Progressive Multi-Scale Exploration.} 
Participants rarely approached exploration with a single precise query. Instead, they followed a progressive strategy, first obtaining an overview of the embedding and then refining their search. Users initially identified clusters corresponding to watch types or visually coherent groups before navigating to local neighborhoods to inspect individual watches and compare design details.
This highlights the importance of supporting multi-scale exploration in visualization systems for large collections. A combination of global overview, smooth navigation through zooming and panning, and linked detail views enables users to move naturally from high-level exploration to detailed inspection, particularly when collections contain more items than can be inspected individually.

\paragraph{Attribute Filtering in Embeddings.} 
While spatial embeddings intuitively represent visual similarity, exploration tasks often involve additional constraints not captured by visual attributes. In our domain, participants frequently combined aesthetic preferences with practical criteria such as price, brand, or case size.
Integrating attribute-based filtering with spatial exploration proved effective for supporting such tasks. Users could first navigate the latent space to discover visually similar watches, then refine the results with metadata filters. This combination supports both exploratory discovery and targeted search. More broadly, embedding-based visualizations benefit from integrating attribute filtering to incorporate non-visual constraints into the analysis.

\paragraph{Search-by-Example in Visual Embeddings.} 
Supporting search-by-example interactions proved valuable. Many exploration tasks begin with a concrete visual reference rather than an abstract query. Allowing users to upload a reference image and insert it into the latent space provides an intuitive entry point for similarity-based exploration.
Positioning the query image within the embedding enables immediate inspection of nearby items with similar visual characteristics. This reduces the cognitive effort required to formulate complex queries and supports more natural exploration. More generally, enabling search-by-example can improve accessibility and usability in visualization systems for visual collections, particularly for non-expert users.

\paragraph{Handling Heterogeneous Attribute Spaces.} 
Our work highlights the challenge of visualizing datasets with heterogeneous attribute spaces, in which attributes differ in dimensionality, distribution, and semantics. As shown in our motivational example (see \autoref{fig:motivationalExample}), naive attribute combination can distort similarity relationships and lead to misleading embeddings.
Our multi-attribute embedding integrates attribute-specific similarity measures into a unified probabilistic model, enabling flexible balancing between heterogeneous attributes while preserving meaningful neighborhood relationships. More generally, embedding techniques for visualization should account for heterogeneous attribute spaces rather than relying on simple attribute concatenation.

\section{Conclusion and Future Work}
\label{Sec:Conclusion}
This paper presents a visualization system to explore the horological design space using a multi-attribute latent representation. Building on UMAP~\cite{McInnes2020}, we addressed the challenge of embedding heterogeneous attribute sets that differ in dimensionality, scale, and semantic meaning. Our approach integrates multiple visual attributes, such as dial color and design, into a unified embedding while incorporating class-aware layout constraints that preserve global semantic structure.

Based on this representation, we developed an interactive visualization system that enables exploration of the latent space through spatial navigation, attribute filtering, and search-by-example interactions. A user study showed that both experts and novices could interpret the embedding, discover related watches, and identify stylistic relationships. Beyond the application domain, the design and evaluation reveal broader implications for visualizing heterogeneous datasets, including balancing global semantic structure with local similarity, supporting progressive multi-scale exploration, and combining spatial embeddings with attribute-based filtering. Although demonstrated in horological design, the system is applicable to other domains in which heterogeneous attributes define similarity relationships.

\setlength{\tabcolsep}{3pt}

 \begin{table}[t]
     \centering
     \caption{
    \revise{
Preliminary comparison with other baseline methods on the watch data set. Error $E$ from Eq.~\eqref{eq:error-metric} is reported for different attribute weights and dimensionality reduction techniques.
} }
     \small
     \begin{tabular}{cc|cccccc|cc}
        \toprule
        \multicolumn{2}{c|}{Weights} 
        & \multicolumn{2}{c}{PCA} 
        & \multicolumn{2}{c}{t-SNE} 
        & \multicolumn{2}{c|}{UMAP} 
        & \multicolumn{2}{c}{Ours} 
        \\
        $w_1$ & $w_2$                
        & Joint & N. joint  & Joint & N. joint & Joint & N. joint  & t-SNE   & UMAP \\
        \midrule
        $0$ & $1$ 
        &  0.8369     & 0.8395          & 0.2465        & 0.2667            & 0.1605        & 0.1607            & 0.2249        &  \textbf{0.1595} \\
        $\nicefrac{1}{4}$ & $\nicefrac{3}{4}$ 
        &  0.4992     & 0.5009          &  0.1326       & 0.1393            & 0.4501        & 0.3510            & 0.0946        &  \textbf{0.0635} \\
        $\nicefrac{1}{2}$ & $\nicefrac{1}{2}$  
        &  0.4148     & 0.4158          &  0.2311       & 0.2438            & 0.3482        & 0.3656            & 0.1366        &  \textbf{0.1220} \\
        $\nicefrac{3}{4}$ & $\nicefrac{1}{4}$ 
        &  0.5803     & 0.5960          &  0.3548       & 0.3368            & 0.4502        & 0.4395            & 0.2136        &  \textbf{0.1744} \\
        $1$ & $0$ 
        &  0.8925     & 0.9287          & 0.2943        & 0.4486            & 0.3464        & 0.3393            &\textbf{0.2834}    &  0.3362 \\
        \bottomrule
     \end{tabular}
     \label{Tab:EvaluationNeighbors2}
 \end{table}
\revise{
Beyond the UMAP-based formulation evaluated in this work, the proposed multi-attribute neighborhood construction can also be transferred to other neighborhood-preserving dimensionality reduction methods. In particular, we implemented a preliminary t-SNE variant in which the high-dimensional affinity matrix is constructed as a weighted combination of attribute-specific affinity matrices, analogous to the weighted graph construction used in our UMAP variant. \autoref{Tab:EvaluationNeighbors2} reports results on the watch data set for standard dimensionality reduction techniques such as PCA, t-SNE, UMAP, as well as our corresponding multi-attribute variants for t-SNE and UMAP. Both of our multi-attribute variants consistently outperform the standard dimensionality-reduction techniques with respect to the metric $E$ in \autoref{eq:error-metric}. While our t-SNE results indicate that the proposed principle is not limited to UMAP, a systematic comparison across different parameter settings, data sets, and evaluation protocols is beyond the scope of this paper and remains an interesting direction for future work.
}

Future work also includes extending the latent representation with temporal or contextual information (e.g., brand evolution or stylistic trends) and integrating multimodal data such as textual descriptions, materials, and mechanical specifications. Adaptive user modeling could further personalize the layout based on individual preferences or browsing behavior. Evaluating the system with broader audiences, including collectors and retailers, will help assess its potential for analysis and recommendation. Overall, this work demonstrates how combining visual embeddings, machine learning, and interactive visualization enables exploration of complex design spaces beyond horology.

\section*{Acknowledgments}
The authors used ChatGPT by OpenAI~\cite{openai_chatgpt_2026} for language editing and drafting assistance. The tool was used to suggest alternative formulations and improve clarity. All scientific claims, methodological choices, experiments, results, figures, and conclusions were reviewed, verified, and finalized by the authors. No AI system was used to generate experimental data or evaluation results.


\bibliographystyle{abbrv-doi-hyperref}

\bibliography{template}

\definecolor{likert1}{RGB}{202,0,32}    
\definecolor{likert2}{RGB}{244,165,130} 
\definecolor{likert3}{RGB}{247,247,247} 
\definecolor{likert4}{RGB}{146,197,222} 
\definecolor{likert5}{RGB}{5,113,176}   

\definecolor{likert71}{RGB}{178,24,43}    
\definecolor{likert72}{RGB}{239,138,98} 
\definecolor{likert73}{RGB}{253,219,199} 
\definecolor{likert74}{RGB}{247,247,247} 
\definecolor{likert75}{RGB}{209,229,240}   
\definecolor{likert76}{RGB}{103,169,207}   
\definecolor{likert77}{RGB}{33,102,172}   


\clearpage

\appendix
\section{Motivational Example}
\label{sec:motivational-example}

\maketitle


To convey the challenges of multi-attribute dimensionality reduction, we illustrate them using a synthetic dataset. Although simplified, this example reflects a common real-world scenario in which objects are described by heterogeneous attributes that differ in dimensionality, statistical distribution, and semantic meaning. Such situations arise in many domains—including product design, fashion, industrial objects, and visual media collections—where items combine categorical properties, aesthetic attributes, and structural characteristics. Integrating these heterogeneous descriptors into a single embedding is essential for enabling meaningful visual exploration and comparison. The dataset used in this example consists of a population of $N$ ellipses, where each object $i$ is characterized by two fundamentally distinct attribute sets.

\paragraph*{Attribute 1: Chromatic Attributes (3D)}
The first attribute is a three-dimensional vector $\mathbf{c} \in [0,1]^3$ encoding the ellipse color in RGB space. To create a clear topological structure, we define two disjoint clusters: a ``hot'' cluster $\mathcal{C}_h$ and a ``cold'' cluster $\mathcal{C}_c$. The color components for each cluster are generated using a uniform distribution $\rchi(min,max)$ as follows:
\vspace{1mm}

\noindent
For the hot cluster $\mathcal{C}_h$:
\begin{align*}
    R_h \sim \rchi(0.8, 1.0), \quad G_h \sim \rchi(0.2, 0.4), \quad B_h \sim \rchi(0.2, 0.4)
\end{align*}
For the cold cluster $\mathcal{C}_c$:
\begin{align*}
    R_c \sim \rchi(0.1, 0.2), \quad G_c \sim \rchi(0.2, 0.4), \quad B_c \sim \rchi(0.7, 1.0)
\end{align*}
In this setup, the color attribute provides a categorical separation between the two groups. 

\paragraph*{Attribute 2: Morphological Attributes (1D)}
The second attribute describes the ellipse's geometric elongation. 
We define the first semi-axis as constant ($a=1$) and vary the second semi-axis $b$ by an elongation factor $r \in \R^+$. 
Unlike the uniform distribution of the color attributes, the elongation is modeled as a Gaussian distribution:

\begin{align*}
    r \sim \mathcal{N}(\mu, \sigma^2) \quad \text{with} \quad \mu=5, \sigma=2
\end{align*}
where $\mathcal{N}(\mu, \sigma^2)$ denotes a normal distribution with mean $\mu$ and variance $\sigma^2$.

This admittedly simple example serves to elucidate the core challenge of multi-attribute data integration. 
Our synthetic dataset comprises $N=400$ ellipses, equally divided into a ``hot'' and a ``cold'' cluster ($n_h=n_c=200$). 
Consequently, each ellipse is represented by a four-dimensional attribute vector $\mathbf{x} \in \R^4$.

The fundamental objective of dimensionality reduction methods, such as UMAP, is to preserve the high-dimensional topological structure in a lower-dimensional representation. 
Specifically, points that are identified as neighbors in the high-dimensional space should remain proximal in the low-dimensional embedding.

To analyze the behavior of these attributes, we first apply UMAP to each attribute space individually. 
As shown in Figure 1 in the main paper, 
when UMAP is applied solely to the three-dimensional color vectors, the bipartite structure of the hot and cold clusters is perfectly preserved. 
Similarly, applying UMAP to the one-dimensional elongation factor--though redundant for a two-dimensional embedding--correctly maintains the continuous Gaussian ordering of the shapes. 
In both isolated cases, the local neighborhood structure is accurately captured.

\noindent
A key question is how to combine these disparate attributes. The simplest approach is attribute concatenation, yielding a joint 4D vector:

\begin{align*}
    \mathbf{x}_{joint} = [w_1\cdot R, w_1\cdot G, w_1\cdot B, w_2\cdot r]^T,
\end{align*}
with weights $w_1,w_2$ to emphasize one attribute.
However, applying UMAP to this concatenated vector fails to preserve the intended neighborhood relationships. 
This failure stems from the fact that the two domains not only follow different statistical distributions (Uniform vs. Gaussian) but also operate on vastly different numerical ranges. 
The elongation factor $r$ (mean $\mu=5$ with high variance) dominates the bounded RGB components ($[0,1]$). 
Consequently, the distance metric is driven by morphology, causing the chromatic structure to vanish in the embedding.
To account for this, we also normalized the attributes using the standard score:
\begin{align*}
    z = \frac{x-\mu}{\sigma}.
\end{align*}
Then we used:
\begin{align*}
    \mathbf{z}_{joint} = [w_1\cdot z_R, w_1\cdot z_G, w_1\cdot z_B, w_2\cdot z_r]^T,
\end{align*}
and applied UMAP as well, see Figure 1 in the main paper. 

To quantify our approach, we introduce a neighborhood preservation metric. For each data point (e.g., an ellipse), we compute the $k$-nearest neighbors in the original attribute spaces: $N_i$ in RGB space (attribute 1) and $N_j$ in morphological space (attribute 2).
After dimensionality reduction with UMAP, we determine the corresponding $k$-nearest neighbors in the 2D embedding, denoted as $N_{\text{UMAP}}$. 
Our system supports combining $m$ attributes $\text{attr}_i$ with weights $w_i$, where:

\begin{align}
w_i \ge 0 \quad \text{and} \quad \sum_{i=1}^{m} w_i = 1
\end{align}
To assess how well the embedding reflects the influence of each attribute, we calculate the Jaccard index $J_i$ between the high-dimensional neighborhood of attribute $i$ and the UMAP neighborhood:
\begin{align}
J_i = \frac{|N_{\text{attr}_i} \cap N_{\text{UMAP}}|}{|N_{\text{attr}_i} \cup N_{\text{UMAP}}|}
\end{align}
We then define the reconstruction error $E$ as the squared deviation between the observed neighborhood overlap and the assigned weights:
\begin{align}
E = \sum_{i=1}^{m} \delta_i\cdot(J_i - w_i)^2,~~~~\delta_i = \left\{
\begin{array}{ll}
0 & J_i \geq w_i \\
1 & \, \textrm{otherwise.} \\
\end{array}
\right.
\label{eq:error-metric}
\end{align}
The correction parameter $\delta_i$ ensures that errors are penalized only if there are not at least $w_i$ common neighbors in $J_i$.

The rationale is as follows: if only one attribute is considered ($w_i = 1$), the ideal embedding perfectly preserves that neighborhood, yielding $J_i = 1$ and $E = 0$. If two attributes are weighted equally ($w_1 = w_2 = 0.5$), the optimal compromise--where the 2D neighborhood contains equal neighbors from both spaces--gives $J_1 = J_2 = 0.5$ and thus $E = 0$.
We report weights and corresponding error metrics for $k=16$ nearest neighbors in~\autoref{Tab:Motivation} for the data in the motivational example in \autoref{fig:motivationalExample} in the main paper. 
This experiment demonstrates that stacking and normalized stacking struggle with neighborhood preservation. In contrast, the method proposed in this paper (ours), achieves a stronger score.

\begin{table}
    \centering
    \caption{Error $E$ from Eq.~\eqref{eq:error-metric} for different weights and techniques.}
    \begin{tabular}{c|ccc}\toprule
       Weights                  & Joint & Normalizing joint  & Our \\\midrule
       $w_1=0.00$, $w_2=1.00$   & 0.01  & 0.01  &  0.01 \\
       $w_1=0.25$, $w_2=0.75$   & 0.46  & 0.31  &  \textbf{0.04} \\
       $w_1=0.50$, $w_2=0.50$   & 0.19  & 0.18  &  \textbf{0.04} \\
       $w_1=0.75$, $w_2=0.25$   & 0.11  & 0.11  &  \textbf{0.05} \\
       $w_1=1.00$, $w_2=0.00$   & 0.01  & 0.01  &  0.01 \\ \bottomrule
    \end{tabular}
    \label{Tab:Motivation}
\end{table}

\section{Questionnaire}

To comprehensively assess the usability and experiential quality of the proposed visualization system, participants completed a post-task questionnaire consisting of standardized and custom items. 
The instrument combined the \textit{System Usability Scale (SUS)}~\cite{brooke1996sus}, the short version of the \textit{User Experience Questionnaire (UEQ-S)}~\cite{hinderks2017}, 
and additional domain-specific questions addressing \textit{Effectiveness} and \textit{Interpretability}.

\paragraph*{Rationale for Using UEQ-S.}
\revise{%
We chose the short version of the UEQ (UEQ-S) instead of the original 26-item instrument to minimize participant fatigue and maintain focus on the most relevant experiential dimensions. 
UEQ-S is a validated, psychometrically sound version of the full UEQ, designed for concise evaluations while preserving sensitivity and reliability. 
It measures two higher-level components of user experience: \textit{Pragmatic Quality} (goal-directed aspects such as efficiency, dependability, and clarity) 
and \textit{Hedonic Quality} (non-task-related aspects such as stimulation, novelty, and enjoyment). 
Each item is rated on a 7-point bipolar semantic differential scale ranging from -3 (most negative) to +3 (most positive).
}

\subsection{Scales Overview.}
\begin{itemize}
    \item \textbf{Demographics:} Questions related to participants’ demographic and background information.
    \item \textbf{System Usability Scale (SUS):} 10 items on a 5-point Likert scale (1 = strongly disagree, 5 = strongly agree). 
    Items alternate between positive and negative wording and are scored according to the standard SUS procedure.
    \item \textbf{User Experience Questionnaire Short (UEQ-S):} 8 bipolar items on a 7-point scale (–3 to +3), measuring \textit{Pragmatic Quality} and \textit{Hedonic Quality}.
    \item \textbf{Custom Questions:} Two sets of additional items (5 each) addressing \textit{Effectiveness}, \textit{Interpretability}, and \textit{Usefulness} 
    rated on the same 5-point Likert scale as SUS to maintain consistency.
\end{itemize}

\subsection{Demographics}

\begin{enumerate}
    \item  Age: \freeresponse[0.1\linewidth] \\[5pt]
    \item  Gender: 
    Female, Male, Non-binary, Prefer not to say  \\[5pt]
    \item  How familiar are you with wristwatches? (1 = Not at all, 7 = Very familiar): \freeresponse[0.1\linewidth] \\[5pt]
    \item  How often do you explore or purchase watches online?:\\[2pt] \freeresponse[0.9\linewidth] \\[5pt]
    \item  Have you used similar interactive visualization systems before? (Yes/No): \freeresponse[0.1\linewidth] \\[5pt]
\end{enumerate}

\subsection{Custom Questions (5-point scale, 1 = strongly disagree, 5 = strongly agree)}
\paragraph*{Effectiveness}
\begin{enumerate}
    \item [E1]The visualization helped me find visually or semantically similar watches easily.\\ \likert  
    \item  [E2] I could efficiently identify watches that matched specific attributes or criteria.  \\ \likert
    \item[E3] The layout supported meaningful comparison between watch types.  \\ \likert
    \item [E4] I was able to complete exploration and comparison tasks without difficulty.  \\ \likert
    \item [E5] The system provided results consistent with my expectations.  \\ \likert
\end{enumerate}

\paragraph*{Interpretability}
\begin{enumerate}
    \item [I1] I understood why certain watches were positioned close to or far from each other.  \\ \likert
    \item [I2] The spatial layout of the latent space made sense to me.  \\ \likert
    \item [I3] The visual cues (e.g., color, position, grouping) helped me interpret relationships between watches.  \\ \likert
    \item [I4] I could infer which attributes (e.g., watch type, color, or design) influenced the visualization.  \\ \likert
    \item [I5] The representation clearly communicated the overall structure of the watch collection. \\ \likert 
\end{enumerate}

\subsection{System Usability Scale (SUS, 5-point scale, 1 = strongly disagree, 5 = strongly agree):}
\begin{enumerate}
    \item [S1] I would like to use this system frequently.  \\ \likert
    \item [S2] I found the system unnecessarily complex. \\ \likert
    \item [S3] I thought the system was easy to use.  \\ \likert
    \item [S4] I think that I would need the support of a technical person to be able to use this system.  \\ \likert
    \item [S5] I found the various functions in this system were well integrated. \\ \likert
    \item [S6] I thought there was too much inconsistency in this system. \\ \likert
    \item [S7] I would imagine that most people would learn to use this system very quickly.  \\ \likert
    \item [S8] I found the system very cumbersome to use.   \\ \likert
    \item [S9] I felt very confident using the system.  \\ \likert
    \item [S10] I needed to learn a lot of things before I could get going with this system.   \\ \likert
\end{enumerate}

\subsection{User Experience Questionnaire Short (UEQ-S, 7-point bipolar scale, –3 to +3):}
\begin{itemize}
    \item \textbf{Pragmatic Quality:}
        \begin{enumerate}
            \item [UP1] obstructive – supportive  \\ \likertSeven
            \item [UP2] complicated – easy  \\ \likertSeven
            \item [UP3] inefficient – efficient  \\ \likertSeven
            \item [UP4] confusing – clear  \\ \likertSeven
        \end{enumerate}
    \item \textbf{Hedonic Quality:}
        \begin{enumerate}
            \item [UH1] boring – exciting  \\ \likertSeven
            \item [UH2] not interesting – interesting  \\ \likertSeven
            \item [UH3] conventional – inventive  \\ \likertSeven
            \item [UH4] usual – leading edge  \\ \likertSeven
        \end{enumerate}
\end{itemize}

\subsection{Scoring and Analysis.}
\revise{%
For the qualitative evaluation, we focused on in-depth analysis of participant feedback rather than formal statistical testing. 
A total of 10 participants (four experts and six novices) took part in the study. 
SUS responses were interpreted following Brooke’s standard scoring procedure (0–100 scale) to provide an overall impression of perceived usability~\cite{brooke1996sus}, 
while UEQ-S ratings were examined descriptively across the two main dimensions (\textit{Pragmatic} and \textit{Hedonic Quality}) according to the procedure described in Hinderks~\cite{hinderks2017}. 
The custom Likert-scale items on Effectiveness and Interpretability, as well as open-ended comments, were analyzed thematically to identify recurring perceptions, usability issues, and differences between expert and novice participants. 
This qualitative approach allowed us to capture nuanced insights into how participants experienced and understood the visualization rather than focusing solely on aggregated numerical scores.
}
The results of the questionnaires can be seen in \autoref{tab:EvaluationResults_Effective}, \autoref{tab:EvaluationResults_Interpretability}, \autoref{tab:EvaluationResults_SUS}, and \autoref{tab:EvaluationResults_UEQ}.

\subsection{Demographics - Results}

\begin{enumerate}
    \item Age (Experts): 40, 47, 57, 67
    \item \revise{Age (Novices): 23, 29, 31, 33, 35, 38}
    \item \revise{Gender (Experts): f, m, m, m}
    \item \revise{Gender (Novices): f, f, m, m, m, m}
    \item \revise{Familiarity with wristwatches (Experts, 1--7): 6, 7, 7, 6}
    \item \revise{Familiarity with wristwatches (Novices, 1--7): 1, 2, 3, 4, 2, 2}
    \item \revise{Online watch exploration frequency (Experts): responses ranged from approximately 3--4 times per week to 4 times per month.}
    \item \revise{Online watch exploration frequency (Novices): most participants reported infrequent exploration, typically only occasionally or when considering a purchase, with the highest frequencies around once per month.}
    \item \revise{Prior use of similar interactive visualization systems (Experts): no, no, no, no}
    \item \revise{Prior use of similar interactive visualization systems (Novices): no, no, no, no, no, no}
\end{enumerate}

\definecolor{Likert1}{RGB}{202,0,32}    
\definecolor{Likert2}{RGB}{244,165,130} 
\definecolor{Likert3}{RGB}{247,247,247} 
\definecolor{Likert4}{RGB}{146,197,222} 
\definecolor{Likert5}{RGB}{5,113,176}   

\newcommand{\likertbox}[1]{%
  \begin{tikzpicture}[baseline=-0.ex,scale=0.5]
    \edef\col{Likert#1}
    \fill[\col,rounded corners=1pt] (0,0) rectangle (0.4,0.4);
    \draw[black!90] (0,0) rectangle (0.4,0.4);
  \end{tikzpicture}%
}

\newcommand{\likertboxSeven}[1]{%
  \begin{tikzpicture}[baseline=-0.ex,scale=0.5]
    \edef\col{likert7#1}
    \fill[\col,rounded corners=1pt] (0,0) rectangle (0.4,0.4);
    \draw[black!90] (0,0) rectangle (0.4,0.4);
  \end{tikzpicture}%
}

\begin{table}[t]
\centering
\begin{tabular}{p{0.8cm}|c|ccccc}
\toprule
 &  & \likertbox{1} & \likertbox{2} & \likertbox{3} & \likertbox{4} & \likertbox{5} \\
\hline
\multirow{2}{*}{E1} 
   & Experts & -- & -- & -- & -- & 4 \\ 
   & Novices & -- & -- & 2 & 2 & 2 \\
\hline
\multirow{2}{*}{E2} 
   & Experts & -- & -- & -- & -- & 4 \\ 
   & Novices & -- & -- & 1 & 3 & 2 \\
\hline
\multirow{2}{*}{E3} 
   & Experts & -- & -- & -- & 1 & 3 \\ 
   & Novices & -- & -- & 2 & 3 & 1 \\
\hline
\multirow{2}{*}{E4} 
   & Experts & -- & -- & -- & 2 & 2 \\ 
   & Novices & -- & -- & 1 & 4 & 1 \\
\hline
\multirow{2}{*}{E5} 
   & Experts & -- & -- & -- & 3 & 1 \\ 
   & Novices & -- & -- & 1 & 3 & 2 \\
\bottomrule
\multicolumn{7}{c}{Strongly disagree~~\likertbox{1}\,\likertbox{2}\,\likertbox{3}\,\likertbox{4}\,\likertbox{5}\, Strongly agree}
\end{tabular}
\caption{Evaluation results for questions regarding the Effectiveness E1-E5.}
\label{tab:EvaluationResults_Effective}
\end{table}

\begin{table}[t]
\centering
\begin{tabular}{p{0.8cm}|c|ccccc}
\toprule
 &  & \likertbox{1} & \likertbox{2} & \likertbox{3} & \likertbox{4} & \likertbox{5} \\
\hline
\multirow{2}{*}{I1} 
   & Experts & -- & -- & -- & 2 & 2 \\ 
   & Novices & -- & -- & 3 & 2 & 1 \\
\hline
\multirow{2}{*}{I2} 
   & Experts & -- & -- & -- & 1 & 3 \\ 
   & Novices & -- & -- & 2 & 2 & 2 \\
\hline
\multirow{2}{*}{I3} 
   & Experts & -- & -- & -- & 3 & 1 \\ 
   & Novices & -- & -- & 1 & 4 & 1 \\
\hline
\multirow{2}{*}{I4} 
   & Experts & -- & -- & -- & 3 & 1 \\ 
   & Novices & -- & -- & 3 & 2 & 1 \\
\hline
\multirow{2}{*}{I5} 
   & Experts & -- & -- & -- & 3 & 1 \\ 
   & Novices & -- & -- & 2 & 2 & 2 \\
\bottomrule
\multicolumn{7}{c}{Strongly disagree~~\likertbox{1}\,\likertbox{2}\,\likertbox{3}\,\likertbox{4}\,\likertbox{5}\, Strongly agree}
\end{tabular}
\caption{Evaluation results for questions regarding the Interpretability I1-I5.}
\label{tab:EvaluationResults_Interpretability}
\end{table}

\begin{table}[t]
\centering
\begin{tabular}{p{0.8cm}|c|ccccc}
\toprule
 &  & \likertbox{1} & \likertbox{2} & \likertbox{3} & \likertbox{4} & \likertbox{5} \\
\hline
\multirow{2}{*}{S1} 
   & Experts & -- & -- & -- & 1 & 3 \\ 
   & Novices & -- & -- & 2 & 3 & 1 \\
\hline
\multirow{2}{*}{S2} 
   & Experts & 1 & 2 & 1 & -- & -- \\ 
   & Novices & 2 & 1 & 2 & 1 & -- \\
\hline
\multirow{2}{*}{S3} 
   & Experts & -- & -- & -- & 1 & 3 \\ 
   & Novices & -- & -- & 3 & 2 & 1 \\
\hline
\multirow{2}{*}{S4} 
   & Experts & 3 & 1 & -- & -- & -- \\ 
   & Novices & -- & 3 & 2 & 1 & -- \\
\hline
\multirow{2}{*}{S5} 
   & Experts & -- & -- & -- & 2 & 2 \\ 
   & Novices & -- & -- & 2 & 3 & 1 \\
\hline
\multirow{2}{*}{S6} 
   & Experts & 3 & 1 & -- & -- & -- \\ 
   & Novices & 1 & 1 & 3 & 1 & -- \\
\hline
\multirow{2}{*}{S7} 
   & Experts & -- & -- & -- & -- & 4 \\ 
   & Novices & -- & -- & 3 & 1 & 2 \\
\hline
\multirow{2}{*}{S8} 
   & Experts & 2 & 2 & -- & -- & -- \\ 
   & Novices & 1 & 3 & 1 & -- & -- \\
\hline
\multirow{2}{*}{S9} 
   & Experts & -- & -- & -- & 2 & 2 \\ 
   & Novices & -- & -- & 2 & 3 & 1 \\
\hline
\multirow{2}{*}{S10} 
   & Experts & 3 & 1 & -- & -- & -- \\ 
   & Novices & -- & 3 & 2 & 1 & -- \\
\bottomrule
\multicolumn{7}{c}{Strongly disagree~~\likertbox{1}\,\likertbox{2}\,\likertbox{3}\,\likertbox{4}\,\likertbox{5}\, Strongly agree}
\end{tabular}
\caption{Evaluation results for SUS questions S1-S10.}
\label{tab:EvaluationResults_SUS}
\end{table}

\begin{table}[t]
\centering
\begin{tabular}{
    p{0.8cm}|
    c|
    p{0.3cm} p{0.3cm} p{0.3cm} p{0.3cm} p{0.3cm} p{0.3cm} p{0.3cm}
}
\toprule
 &  & \likertboxSeven{1} & \likertboxSeven{2} & \likertboxSeven{3} 
   & \likertboxSeven{4} & \likertboxSeven{5} & \likertboxSeven{6} & \likertboxSeven{7} \\
\hline

\multirow{2}{*}{UP1}
   & Experts & -- & -- & -- & -- & -- & 2 & 2 \\
   & Novices & -- & -- & -- & -- & 2 & 2 & 2 \\
\hline

\multirow{2}{*}{UP2}
   & Experts & -- & -- & -- & -- & -- & 1 & 3 \\
   & Novices & -- & -- & 1 &2 & 2 & 1 & -- \\
\hline

\multirow{2}{*}{UP3}
   & Experts & -- & -- & -- & -- & -- & 3 & 1 \\
   & Novices & -- & -- & -- & 1 & 1 & 2 & 2 \\
\hline

\multirow{2}{*}{UP4}
   & Experts & -- & -- & -- & -- & -- & 3 & 1 \\
   & Novices & -- & -- & -- & 2 & 1 & 2 & 1 \\
\hline

\multirow{2}{*}{UH1}
   & Experts & -- & -- & -- & -- & -- & 2 & 2 \\
   & Novices & -- & -- & -- & -- & 1 & 3 & 2 \\
\hline

\multirow{2}{*}{UH2}
   & Experts & -- & -- & -- & -- & -- & -- & 4 \\
   & Novices & -- & -- & -- & -- & -- & 3 & 3 \\
\hline

\multirow{2}{*}{UH3}
   & Experts & -- & -- & -- & -- & -- & 1 & 3 \\
   & Novices & -- & -- & -- & -- & 2 & 3 & 1 \\
\hline

\multirow{2}{*}{UH4}
   & Experts & -- & -- & -- & -- & -- & 1 & 3 \\
   & Novices & -- & -- & -- & 2 & 1 & 2 & 1 \\
\bottomrule

\multicolumn{9}{c}{
Strongly disagree~~
\likertboxSeven{1}\,\likertboxSeven{2}\,\likertboxSeven{3}\,\likertboxSeven{4}\,\likertboxSeven{5}\,\likertboxSeven{6}\,\likertboxSeven{7}\,
Strongly agree
}

\end{tabular}
\caption{Evaluation results for UEQ-S questions regarding Pragmatic Quality (UP1-UP4) and Hedonic Quality (UH1-UH4).}
\label{tab:EvaluationResults_UEQ}
\end{table}

\end{document}